\documentclass[runningheads]{llncs}

\usepackage{eccv}

\usepackage{eccvabbrv}

\usepackage{graphicx}
\usepackage{booktabs}

\usepackage[accsupp]{axessibility}  % Improves PDF readability for those with disabilities.

\usepackage{amsmath}
\usepackage{amssymb}
\usepackage{mathtools}
\usepackage{algorithm}
\usepackage{algorithmic}

\usepackage{url}
\usepackage{enumitem}
\usepackage{colortbl}
\usepackage{xcolor}
\usepackage{multirow}
\usepackage{makecell}
\usepackage{wrapfig}
\usepackage{bbding}

\usepackage[table,dvipsnames]{xcolor}
\definecolor{rowgray}{gray}{0.92}
\definecolor{mypink}{RGB}{218,0,130}
\definecolor{Lavender}{RGB}{230,230,250}
\definecolor{LightPink}{RGB}{255,182,193}
\definecolor{myred}{RGB}{250,220,220}
\definecolor{myyellow}{RGB}{255, 242, 204}

\usepackage{hyperref}

\usepackage{orcidlink}

\begin{document}

\title{LaGen: Towards Autoregressive LiDAR Scene Generation} 

\titlerunning{LaGen}

\author{Sizhuo Zhou\inst{1,2} \orcidlink{0009-0001-1147-4469} \and
Xiaosong Jia\inst{4}\textsuperscript{*} \orcidlink{0000-0002-5222-1476} \and
Fanrui Zhang\inst{1,2} \orcidlink{0000-0002-1078-430X} \and
Junjie Li\inst{1,2} \orcidlink{0009-0008-9116-3986} \and 
Juyong Zhang\inst{1} \orcidlink{0000-0002-1805-1426} \and
Yukang Feng\inst{2} \orcidlink{0009-0000-5594-3629} \and
Jianwen Sun\inst{2} \orcidlink{0009-0000-3145-5225} \and
Songbur Wong\inst{3} \and
Junqi You\inst{3} \and
Junchi Yan\inst{2,3}\textsuperscript{*} \orcidlink{0000-0001-9639-7679}}

\authorrunning{S. Zhou et al.}

\institute{University of Science and Technology of China, Hefei, Anhui 230026, China \and
Shanghai Innovation Institute, Shanghai 200231, China \and 
Shanghai Jiao Tong University, Shanghai 200240, China \and
Fudan University, Shanghai 200433, China \\
\email{yanjunchi@sjtu.edu.cn}
}

\begingroup \renewcommand{\thefootnote}{*} \footnotetext[1]{Corresponding author.} \endgroup

\maketitle

%--------------------------------------------------------------------
\begin{figure}[ht]
  \centering
  \includegraphics[width=\linewidth]{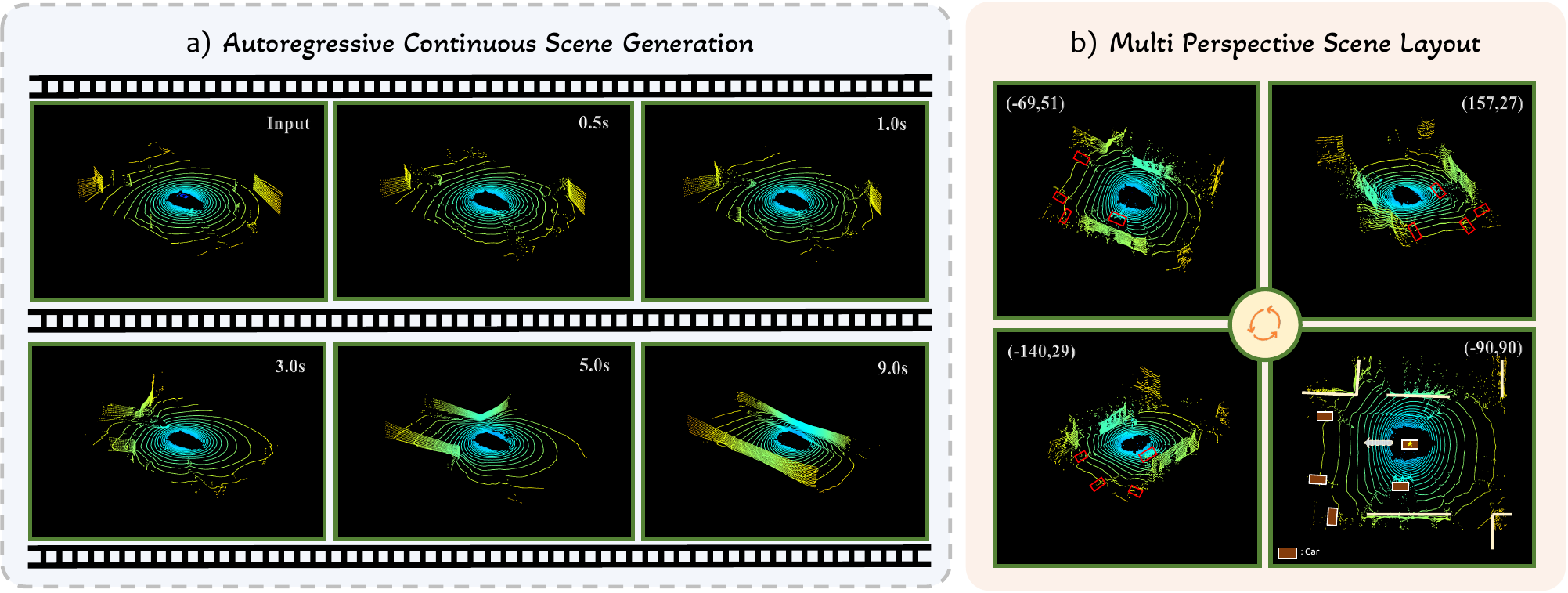}\\
  \caption{\label{teaser} 
    This work introduces LaGen, a novel 4D LiDAR scene generation framework that can autoregressively generate \textbf{(a)} long-horizon autonomous driving scenes based solely on single-frame input. It is capable of generating high-fidelity LiDAR consistent with the real world; \textbf{(b)} illustrates the visualization results from different viewpoints.}
\end{figure}
%--------------------------------------------------------------------

\begin{abstract}
Generative world models for autonomous driving (AD) are of great value in applications such as data augmentation, closed-loop simulation, and safety-critical scenario evaluation.
Unlike the widely studied image modality, in this work we explore generative world models for LiDAR data.
Existing generation methods for LiDAR predominantly focus on single frame generation or lack the capacity for interactive simulation, while existing prediction approaches require multiple frames of historical input and can only deterministically predict multiple frames at once.
Both paradigms fail to support long-horizon interactive generation.
To this end, we introduce \textbf{LaGen}, which, to the best of our knowledge is the first autoregressive framework capable of generating long-horizon LiDAR scenes in a frame-by-frame, interactive manner.
LaGen is able to take a single-frame input as a starting point and effectively utilize bounding box information as conditions to generate high-fidelity 4D scene.
In addition, we introduce a scene decoupling estimation module to enhance the model's interactive generation capability for object-level content, as well as a noise modulation module to mitigate error accumulation during long-horizon generation.
We extensively evaluate LaGen's performance in controlled data generation and long-horizon scene generation on the nuScenes dataset.
The experimental results demonstrate that LaGen achieves state-of-the-art performance, especially on later frames.
The code is publicly available at: https://github.com/szzhou88/LaGen.
\end{abstract}

\section{Introduction}
\label{sec:intro}

In recent years, autonomous systems have attracted increasing attention in both academia and industry. 
LiDAR is one of the key sensors in various autonomous systems. 
It provides precise and rich geometric information about the surrounding environment \cite{li2024di, li2023lwsis, SECOND, cheng2023language, yin2024fusion} and has been widely applied across a range of fields \cite{shi2020pv, shi2019pointrcnn,malavazi2018lidar, weiss2011plant, resop2019drone, zhang2014loam}.
Unlike the extensively studied image modality, research on LiDAR is still limited.
Existing studies have focused on two main directions: one is generating high-quality LiDAR data \cite{LiDARGen2022, xiong2023ultralidar, nakashima2024lidar} to reduce the high cost of real-world acquisition \cite{meng2020weakly, wu2020deep}, and the other is predicting LiDAR data \cite{weng2021inverting, weng2022s2net, khurana2023point} to support decision-making in autonomous systems.

However, these research directions encounter substantial limitations in practical applications.
Most existing works on LiDAR generation predominantly focus on learning the data distribution within a dataset, enabling them to synthesize only non-sequential, single LiDAR point clouds. 
Recent work has explored the generation of 4D LiDAR scenes ~\cite{li2025uniscene, ni2025maskgwm, liang2026lidarcrafter}, but it lacks interactivity, is not suitable for long-horizon tasks, and cannot be used for closed-loop autonomous driving simulation.
For LiDAR prediction tasks, with the rise of end-to-end autonomous driving (E2E-AD) \cite{hu2023planning, jiang2023vad, weng2024drive}, current methods are confronted with the following fundamental challenges:
\begin{enumerate}[label=\arabic*), leftmargin=*, align=left]
\item Prediction requires multiple frames of historical scene data; however, in closed-loop simulation \cite{chitta2022transfuser, dosovitskiy2017carla, jia2024bench2drive, prakash2021multi}, only the initial frame is available;
\item These methods are limited to predicting LiDAR scenes along a predetermined trajectory, while the ego vehicle may generate varying trajectories based on different decision-making.
\item They cannot accurately predict LiDAR over long temporal horizons, as they are limited by fixed historical inputs.
\end{enumerate} 

To address these challenges, we explore LiDAR-based generative world models and propose LaGen. 
LaGen is an autoregressive generation framework capable of producing high-fidelity, long-horizon LiDAR scenes on a frame-by-frame basis (see Figure~\ref{teaser}).
Specifically, we first utilize spherical projection to convert the three-dimensional LiDAR data into a more compact form: a range image.
Next, we construct a LiDAR data generator based on the Latent Diffusion Model (LDM) \cite{rombach2022high}, and guide the diffusion process using multiple control conditions.

To further enhance the spatiotemporal consistency of the entire generative framework, we also introduce the following carefully designed strategies:

\begin{enumerate}[label=\arabic*), leftmargin=*, align=left]
\item To improve the spatial consistency, we introduce a Scene Decoupling Estimation (SDE) module. 
It decouples the LiDAR scene into foreground (static and dynamic objects) and background, and estimates the foreground and background of the current frame based on the currentscene state information.
This provides the model with object-level estimations of current scene, thereby enhancing the model's interactive generation capability.
\item To enhance the temporal consistency, we introduce a Noise Modulation (NM) module designed to alleviate error accumulation. 
This is because the autoregressive paradigm suffers from a train-inference mismatch; during inference, errors in the generated samples accumulate over longer temporal horizons.
The proposed module injects noise into the conditional features containing information from the previous frame, thereby reducing the model's over-reliance on prior frames and enabling it to better adapt to imperfect conditions.
\end{enumerate}

We compare LaGen with state-of-the-art LiDAR generation models on the nuScenes dataset.
The experimental results show that our method has clear advantages in generating high-fidelity LiDAR point clouds and temporally consistent LiDAR scenes over long horizons.
Furthermore, we apply LaGen to downstream prediction tasks to further validate its performance in long-horizon generation.
Experimental results indicate that LaGen can accurately generate long-horizon LiDAR scenes using only a single-frame historical input.
It is noteworthy that our autoregressive framework further enables autonomous systems to incorporate decisions made at each time step into the generation of future predictions, thereby naturally supporting interactive world simulation.

In summary, our contributions are as follows:
\begin{itemize}[leftmargin=*]
\item We propose LaGen, which is the first autoregressive framework capable of generating high-fidelity, long-horizon LiDAR scenes with interactivity;
\item We meticulously design a control condition feature encoder to guide the generation process, and propose a scene decoupling estimation module and a noise modulation module, which effectively enhance the model's performance in interactive scene-detail generation and long-horizon autoregressive generation;
\item We thoroughly evaluate LaGen's performance in generating high-fidelity LiDAR data and temporally consistent 4D LiDAR scenes on the nuScenes dataset to demonstrate the advantages of our method, and further assess its performance on long-horizon prediction tasks.
\end{itemize}

\section{Related Work}

%--------------------------------------------------------------------
\begin{figure}[ht]
  \centering
  \includegraphics[width=\linewidth]{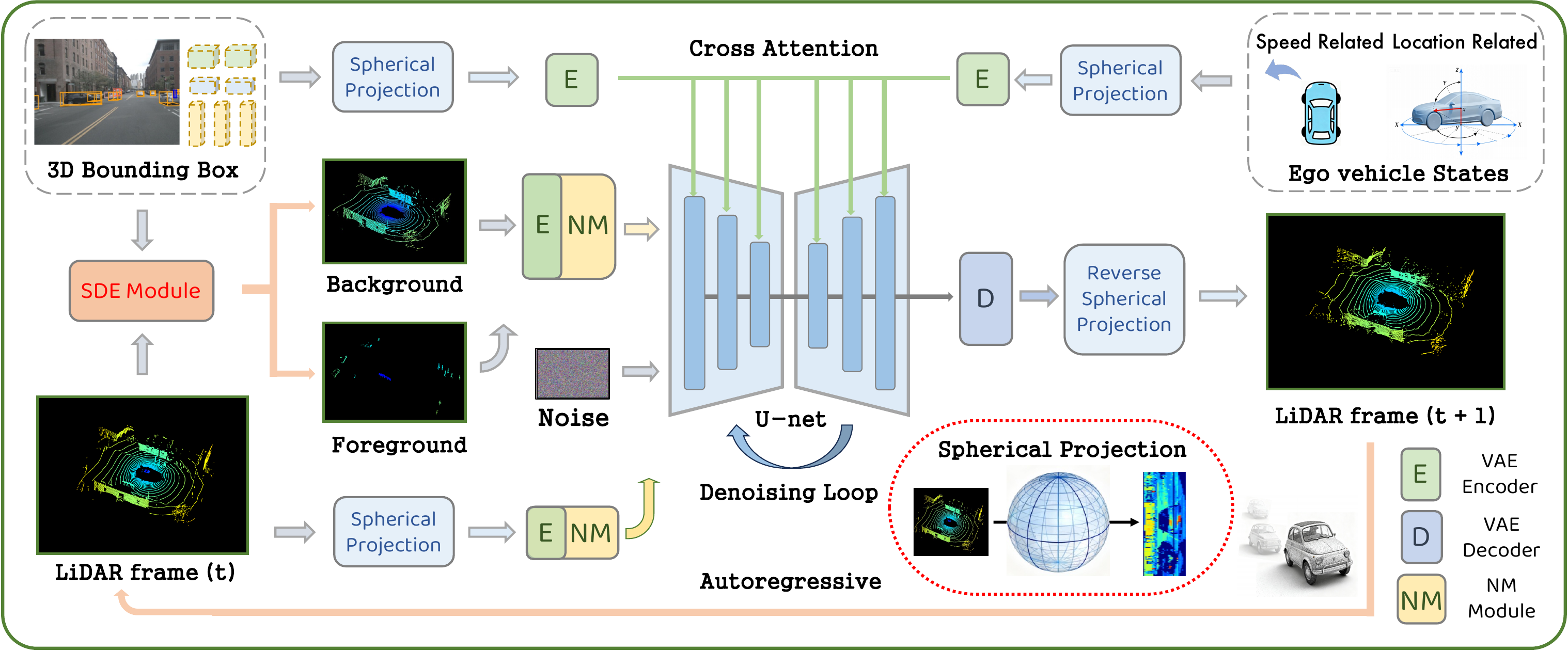}\\
  \caption{\label{overview} 
    \textbf{Overview of the LaGen framework.} It first obtains estimates of the current frame's foreground and background point clouds via the Scene Decoupling Estimation (SDE) module. Subsequently, all three-dimensional information is projected to the two-dimensional range image space via spherical mapping. These range-view representations are then processed by a pretrained VAE and a noise modulation (NM) module, before being passed to the U-net network to generate the LiDAR scene for the current frame. Afterward, the current generated scene is iteratively used as the input to the model, thereby completing the autoregressive process.}
  \end{figure}
%--------------------------------------------------------------------

\noindent\textbf{Generative Models.}
Generative models aim to learn the underlying data distribution and generate novel samples. 
Generative Adversarial Networks (GANs) \cite{goodfellow2014generative} are a prominent example. 
GANs alternately train a pair of competing networks, the generator and the discriminator, to minimize the adversarial objective.
However, GANs often suffer from unstable training dynamics.
Diffusion models (DMs) \cite{sohl2015deep} have shown outstanding performance in the field of image synthesis. 
Mainstream diffusion models include score-based models \cite{song2019generative, song2020improved, song2020score} and Denoising Diffusion Probabilistic Models (DDPM) \cite{ho2020denoising, nichol2021improved, kingma2021variational, saharia2022photorealistic}.
They enable stable training using simple objective functions, addressing challenges associated with GANs.
Recently, Latent Diffusion Models (LDMs) \cite{rombach2022high} have been proposed. 
They perform the diffusion process in the latent space, greatly improving generation efficiency.
They have also enabled significant advances in multimodal conditional generation \cite{nichol2021glide, ramesh2022hierarchical, zhang2023adding}.

\noindent\textbf{LiDAR Data Generation.}
Generating LiDAR point clouds is a 3D data generation task. 
Caccia et al. \cite{caccia2019deep} were the first to represent LiDAR data as range images and employed Variational Autoencoders (VAEs) \cite{kingma2013auto} and Generative Adversarial Networks (GANs) \cite{goodfellow2014generative} for generation.
Subsequently, LiDARGen \cite{LiDARGen2022} introduced a score-based LiDAR data generation model. 
UltraLiDAR \cite{xiong2023ultralidar} voxelizes LiDAR point clouds and employs VQ-VAE \cite{yu2021vector} for generation.
LiDM \cite{ran2024towards} separately generates scenes, objects, and trajectories, and simulates ray casting to produce point clouds.
R2DM \cite{nakashima2024lidar}, based on Denoising Diffusion Probabilistic Models (DDPMs) \cite{song2020denoising}, further improves the fidelity of generated point clouds.
%OLiDM \cite{yan2025olidm} proposes two modules: the Object-Scene Progressive Generation module and the Object Semantic Alignment module, to enhance model performance on downstream perception tasks. 
RangeLDM \cite{hu2024rangeldm} introduces a range-guided discriminator, enabling fast generation through a latent diffusion model.
La La LiDAR \cite{liu2026lidar} introduces a large-scale, layout-guided LiDAR generation model.
However, most of these frameworks are only capable of generating isolated single-frame LiDAR data.
Genesis \cite{guo2026genesis} introduces a unified framework for the joint generation of driving videos and LiDAR sequences. 
DriveLiDAR4D \cite{cai2026drivelidar4d} incorporates three modalities of conditional inputs, namely road sketches, scene descriptions, and object priors, which further enhance the quality of the generated results.
UniScene \cite{li2025uniscene} leverages occupancy as an intermediate representation to generate continuous LiDAR data, thereby reducing the complexity of scene synthesis.
OpenDWM \cite{ni2025maskgwm} is built upon the DiT (Diffusion Transformer) architecture for video generation and has been extended to the domain of LiDAR scenes.
LiDARCrafter \cite{liang2026lidarcrafter}, parses natural language inputs into scene graphs to guide the generation process. 
It also explores the autoregressive genration, but it generates future trajectories and control conditions in a one-shot manner based on initial semantics, lacking interactivity and cannot be used for closed-loop simulation.
In addition, LaGen is primarily based on bounding box conditions, without utilizing natural language information.

\noindent\textbf{LiDAR Data Forecasting.}
Point cloud prediction \cite{mersch2022self, weng2021inverting, weng2022s2net, wilson2023argoverse} is a fundamental self-supervised task in autonomous driving, where multiple frames of historical point clouds are used as input to predict future point clouds. 
Early work predicted future point cloud sequences by predicting scene flow \cite{liu2019flownet3d, zhai2021optical}.
Recent studies have started to directly predict from raw point cloud data \cite{luo2023pcpnet, pal2024atppnet, weng2021inverting, weng2022s2net}. 
4D-Occ \cite{khurana2023point} predicts point clouds from the perspective of geometric occupancy prediction, which further improves prediction accuracy.
ViDAR \cite{yang2024visual} focuses on visual point cloud prediction, using historical images to forecast future point clouds, which improves its performance in downstream applications to some extent.
These frameworks typically require multiple frames of historical input, yet still cannot accurately predict long-horizon LiDAR scenes and lack interactivity.

\section{Method}
\subsection{Preliminaries}
\noindent\textbf{Range Image: A Compact Representation for LiDAR Data.}
Range images offer a compact representation of LiDAR data by parametrizing unstructured point clouds in 3D space into a 2D pixel space. 
Specifically, for a point \( p = (x, y, z) \in \mathbb{R}^3 \) in Cartesian coordinates, it is transformed to the spherical coordinates \( (r, \theta, \phi) \) via spherical projection.
We adopt the corrected range image representation \cite{hu2024rangeldm} as follows:
\begin{equation}
\label{corrected_range}
\begin{aligned}
    r &= \sqrt{x^{2} + y^{2} + (z - h_{j})^{2}}, \\
    \theta &= \arctan(y, x), \quad \phi = \phi_{j},
\end{aligned}
\end{equation}
where \( r \) denotes the depth value, \( \theta \) is the azimuth angle, \( \phi \) is the elevation angle, \( h_{j} \) and \( \phi_{j} \) denote the sensor height and elevation angle estimated by Hough voting.
Next, we convert the spherical coordinates \( (r, \theta, \phi) \) into pixel coordinates \( (u, v) \) of the range image using the equations:
\begin{equation}
\label{rasterization_2d}
u = W - ((\theta + \pi) / 2\pi) \cdot W, \quad v = H - j,
\end{equation}
where \( W \) and \( H \) are the width and height of the image in pixels.
Finally, we normalize the depth and intensity values of the point cloud to obtain a two-channel range image \( I \).

%--------------------------------------------------------------
\noindent\textbf{Latent Diffusion Model for LiDAR Generation.}
We employ the Latent Diffusion Model \cite{rombach2022high} for LiDAR data generation. 
LDM consists of two components: a VAE and a 2D UNet.
The VAE encodes the input range image \( I \in \mathbb{R}^{H \times W \times 2} \) into a latent representation \( \hat{z} \in \mathbb{R}^{h \times w \times c} \). 
The decoder then takes latent \( \hat{z} \) as input and generates the reconstructed range image \( \hat{I} \in \mathbb{R}^{H \times W \times 2} \).

During the diffusion process, noise is progressively added to the initial latent \( z_{0} \) to obtain noisy latent \( z_{t} \), with the noise level increasing as the timestep \( t \in \{1, 2, \ldots, T\} \) advances.
The UNet is trained to predict the noise added to the latent feature \( \hat{z} \).
The training loss of the LDM is defined as:
\begin{equation}
\label{ldm_loss}
\mathcal{L}_{\text{LDM}}=\mathbb{E}_{\epsilon_{t} \in \mathcal{N}(0,1), t \in \mathcal{U}[0, T], c}\left[\left\| \epsilon_{t}-\epsilon_{\theta}\left(z_{t} ; t, c\right)\right\| ^{2}\right],
\end{equation}
where \( z_{t} \) denotes the noisy latent at timestep \( t \), \( \epsilon_{\theta} \) is the noise prediction network to be trained, and \( c \) is the encoded conditional information.
During inference, LDM generates range images by iteratively removing the noise predicted by UNet from randomly sampled Gaussian noise over \( T \) steps.

%--------------------------------------------------------------
\subsection{Overall Framework}
Given the first frame of LiDAR data for a driving scenario \( P^{0} \in \mathbb{R}^{N^{0} \times 4} \), our objective is to recursively generate subsequent LiDAR frames \( P^{s} \), for \( s = 1, 2, \ldots, S \), in accordance with the actual motion of the ego-vehicle.

%------------------------------------------------------------------
\begin{wrapfigure}[17]{r}{0.50\textwidth}
    \centering
    \includegraphics[width=\linewidth]{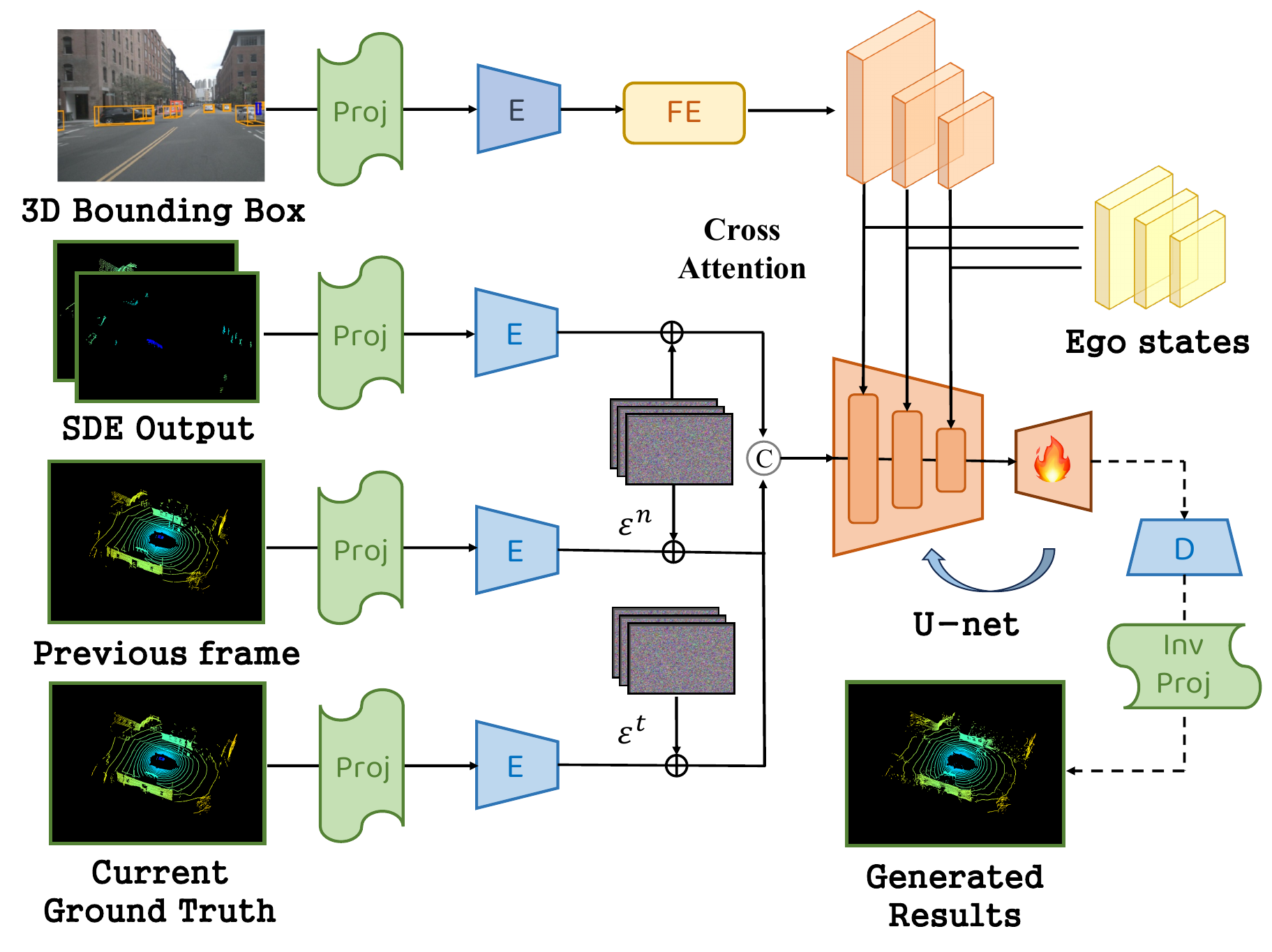}
    \caption{\label{pipeline}
    The main architecture of the high-quality LiDAR data generator.}
\end{wrapfigure}
%------------------------------------------------------------------

In autonomous driving simulators, the following information for each frame of the driving scenario can be tracked \cite{karnchanachari2024towards, jia2024bench2drive}:
\begin{itemize}[leftmargin=*]
\item 3D bounding boxes and semantic labels of the previous and current frames: \( B^{s} = \{(b_{i}, c_{i})\}_{i = 1}^{N_{b}} \), where \( b_{i} = (x_{j}, y_{j}, z_{j})_{j = 1}^{8} \in \mathbb{R}^{8 \times 3} \) denotes the bounding boxes of dynamic and static objects (such as vehicles, pedestrians, and obstacles) within a predefined range, and \( c_{i} \in C_{box} \) is the associated semantic category.
\item Ego states: \( E^{s} = \{E_{\text{ego}},E_{\text{trans}}\} \), where \( E_{\text{ego}} \) includes ego-vehicle speed, acceleration and steering angle information, while \( E_{\text{trans}} \) is the transformation matrix from the ego-vehicle to the global coordinate frame.
%\end{enumerate} 
\end{itemize} 
It should be noted that the bounding boxes of the current frame are not future information; rather, they are produced by the simulator based on the action from the previous frame.
Accordingly, at each step of the generation process, we aim to generate the current LiDAR frame based solely on the previous frame, while incorporating 3D bounding box information and ego-vehicle states. 
The pipeline of LaGen is shown in Figure~\ref{overview}.

%--------------------------------------------------------------
\subsection{Strongly Spatiotemporally Consistent LiDAR Scene Generator and Inference Framework}
In this section, we present the design details of the generator and inference framework for LaGen. 
We construct the core generative architecture based on LDMs and implement conditional control via meticulously designed scene information encoders (see Figure~\ref{pipeline}). 
Furthermore, two crucial modules are introduced to bolster the spatial-temporal consistency of the overall framework.

%--------------------------------------------------------------
\subsubsection{3.3.1. Multimodal Conditional Feature Encoder} \mbox{}\\

\noindent\textbf{Previous Frame's LiDAR Data.}
Using the previous frame's LiDAR point cloud \( P^{s-1} \in \mathbb{R}^{N^{s-1} \times 4} \) as one of the control conditions in the generation process is crucial for endowing the generator with autoregressive inference capability.
To incorporate previous frame information into the denoising process, we first convert the 3D point cloud \( P^{s-1} \) into a 2D range image \( I^{s-1} \in \mathbb{R}^{2 \times H \times W} \) using Equation~(\ref{corrected_range}). The variational autoencoder then projects \( I^{s-1} \) into the latent space as a latent variable \( z^{s-1} \in \mathbb{R}^{c \times h \times w} \).

We can perform the following computation:
\begin{equation}
\label{trans_metrix}
E_{\text{rel}}^{s-1} = (E_{\text{ego2glb}}^{s} \cdot E_{\text{li2ego}}^{s})^{-1} \cdot (E_{\text{ego2glb}}^{s-1} \cdot E_{\text{li2ego}}^{s-1}),
\end{equation}
to obtain the transformation matrix \( E_{\text{rel}}^{s-1} \in \mathbb{R}^{4 \times 4} \), which represents the coordinate transformation from the previous frame's LiDAR sensor to the current frame's LiDAR sensor.
Here, \( E_{\text{ego2glb}}^{s-1} \) and \( E_{\text{ego2glb}}^{s} \) are the transformation matrices from the ego-vehicle to the global coordinate system, while \( E_{\text{li2ego}}^{s-1} \), \( E_{\text{li2ego}}^{s} \) denote the transformation matrices from the LiDAR sensor to the ego-vehicle.
Next, we apply a Feature-wise Linear Modulation (FiLM) \cite{perez2018film} layer to adjust the latent feature \( z^{s-1} \) of the previous frame's LiDAR dat:
\begin{equation}
\label{film}
\hat{z}^{s-1} = FiLM(z^{s-1}, MLP(Flatten(E_{relative}^{s-1}))),
\end{equation}
and inject it into the latent diffusion process of the model.

%--------------------------------------------------------------
\noindent\textbf{Object-Level 3D Bounding Box Projection.}
3D bounding box information is one of the most important cues in autonomous driving scenarios.
To fully exploit this information, we first extract the four corner vertices of \( b_{i} \) that are closest to the ground and perform pairwise interpolation between these vertices to obtain a point cloud \( \hat{b}_{i} \) that encloses the object region.
Next, according to the semantic label \( c_{i} \), we using Equation~(\ref{corrected_range}) to project point clouds of different classes into range image binary masks \( M_{B} \in \mathbb{R}^{D \times H \times W} \), where \( D \) denotes the number of categories.
Then, we encode these masks into latent features \( H_{B} \in \mathbb{R}^{N \times D} \) in the latent space using an object-level mask encoder \( \mathcal{E}_{mask} \).

Subsequently, we inject \( H_{B} \) into multiple layers of the UNet via a cross-attention mechanism:
\begin{equation}
\label{cross_attention}
q = \hat{H}, \quad k = H_{B}^{\text{prev}},\quad v = H_{B}^{\text{cur}},
\end{equation}
where \( q \), \( k \), \( v \) denote the query, key, and value in the cross-attention layer, \( \hat{H} \) represents the features of the intermediate layer, and \( H_{B}^{\text{cur}} \), \( H_{B}^{\text{prev}} \) are the features of bounding box projections for the current and previous frames.

%--------------------------------------------------------------
\noindent\textbf{Ego States.}
We first encode the ego vehicle state \( E_{\text{ego}} \) and \( E_{\text{trans}} \) into compact vector representations. 
These representations are then embedded via a timestep embedding module to capture both temporal and geometric transition information. 
After a nonlinear projection, the resulting embedding is converted into a single conditional token, which is injected into each level of the U-Net via a cross-attention mechanism.

%--------------------------------------------------------------
\subsubsection{3.3.2. Scene Decoupling Estimation} 
\begin{wrapfigure}[18]{r}{0.50\textwidth}
    \centering
    \includegraphics[width=\linewidth]{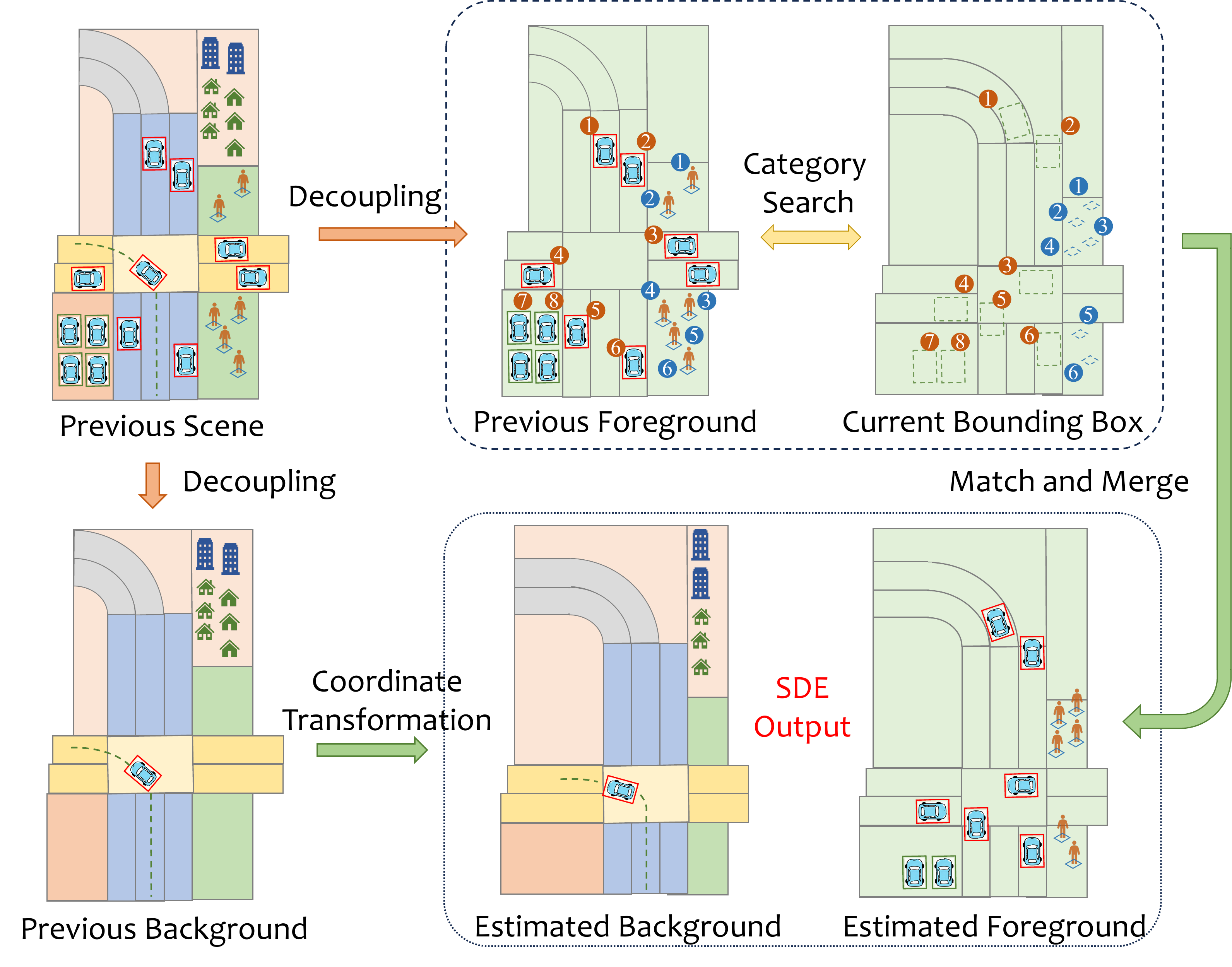}
    \caption{\label{sde_model}
    The schematic diagram of scene decoupling estimation module.}
\end{wrapfigure}

\mbox{}\\
Having established the basic framework for LiDAR data generation conditioned on multiple control conditions, we further design a Scene Decoupling Estimation (SDE) module to improve the quality of generated details (see Figure~\ref{sde_model}).
The design is described in detail below.

For the previous frame LiDAR point cloud \( P^{s-1} \in \mathbb{R}^{N^{s-1} \times 4} \), we decouple it into foreground and background components based on its associated bounding box information \( B^{s-1} \). 
The foreground point cloud \( P^{s-1}_{\text{obj}} \) consists of all points in \( P^{s-1} \) that fall within any of the bounding boxes:
\begin{equation}
\label{foreground_estimate}
P^{s-1}_{\text{obj}}=\{p_{ij}\},i \in\{1,2,...,D\}, j \in\{1,2,...,K_i\},
\end{equation}
%------------------------------------------------------------------
%------------------------------------------------------------------
where \( p_{ij} \) denotes the set of points within the \( j \)-th bounding box of the \( i \)-th category, \( D \) is the total number of semantic categories of bounding boxes, and \( K_i \) is the total number of bounding boxes in category \( i \). 
The background point cloud \( P^{s-1}_{\text{bg}} \) is defined as the set of all points in \( P^{s-1} \) that lie outside the bounding boxes.
For each \( p_{ij}^{s-1} \), we first adjust the viewpoint of the point cloud according to the coordinate transformation matrix of the sensor between the two frames:
\begin{equation}
\label{trans_pij}
\hat{p}_{ij}^{s-1} = trans(P_{ij}^{s-1}, E_{\text{rel}}^{s-1} ),
\end{equation}
To estimate the position of \( \hat{p}_{ij} \) in the current-frame scene, we first compute the centers \( C_{ij}^{s-1} \) and \( C_{ij}^{s} \) of all bounding boxes in the two adjacent frames. 
% For a bounding box \( b_{ij}^{s} \) in the current scene, we search within the same category in the previous frame for the bounding box whose center is closest to \( C_{ij}^{s} \), denoted as \( b_{ij'}^{s-1} \). 
For each bounding box \( b_{ij}^{s} \) in the current scene, we search among the bounding boxes of the same semantic category in the previous frame for the one whose center is nearest to \( C_{ij}^{s} \), denoted as \( b_{ij'}^{s-1} \).
%Thus, we obtain an estimate for each foreground object point cloud \( p_{ij}^{s} \):
Thus, we can obtain an estimate for each foreground object point cloud \( p_{ij}^{s} \):
\begin{equation}
\label{estimation_foreground}
 \widetilde{p}_{ij}^{s} = \hat{p}_{ij}^{s-1} + (C_{ij}^{s} - C_{ij'}^{s-1}), \quad i \in\{1,2,...,D\}.
\end{equation}
Then, we aggregate all \( \widetilde{p}_{ij}^{s} \) to obtain an estimate \( \widetilde{P}^{s}_{\text{obj}} \) of the foreground point cloud \( P^{s}_{\text{obj}} \) in the current scene.

As the background estimation is performed in the ego-centric LiDAR coordinate system, the background point cloud is estimated as follows:

\begin{equation}
\label{estimation_background}
\widetilde{P}^{s}_{\text{bg}} = trans(P^{s-1}_{\text{bg}}, E_{\text{rotate}}^{s-1}).
\end{equation}
% Finally, we project \( \widetilde{P}^{s}_{\text{obj}} \) and \( \widetilde{P}^{s}_{\text{bg}} \) into the range-view space, and after encoding them with the VAE, inject the resulting features into the diffusion process. 
Finally, we project both \( \widetilde{P}^{s}_{\text{obj}} \) and \( \widetilde{P}^{s}_{\text{bg}} \) into the range-view space and encode them using the VAE, after which the resulting features are injected into the diffusion process.

Compared with directly providing bounding box location information, the SDE module better captures object-level details.
Mask information from bounding boxes only provides the model with a coarse understanding of each object's location within the scene, whereas SDE enables the model to better capture the detailed point distribution of each object.
For a more detailed theoretical analysis of this module, please refer to the supplementary materials.

%--------------------------------------------------------------
\subsubsection{3.3.3. Noise Modulation Module and Error Accumulation Mitigation} \mbox{}\\
Autoregressive generation, which relies on information
from the previous frame, often leads to a train-inference discrepancy. 
Specifically, during training, the previous data is always ground truth, whereas during inference, it is generated by the model and inevitably contains some error compared to the real data. 
%------------------------------------------------------------------
\begin{wrapfigure}[15]{r}{0.50\textwidth}
    \centering
    \includegraphics[width=\linewidth]{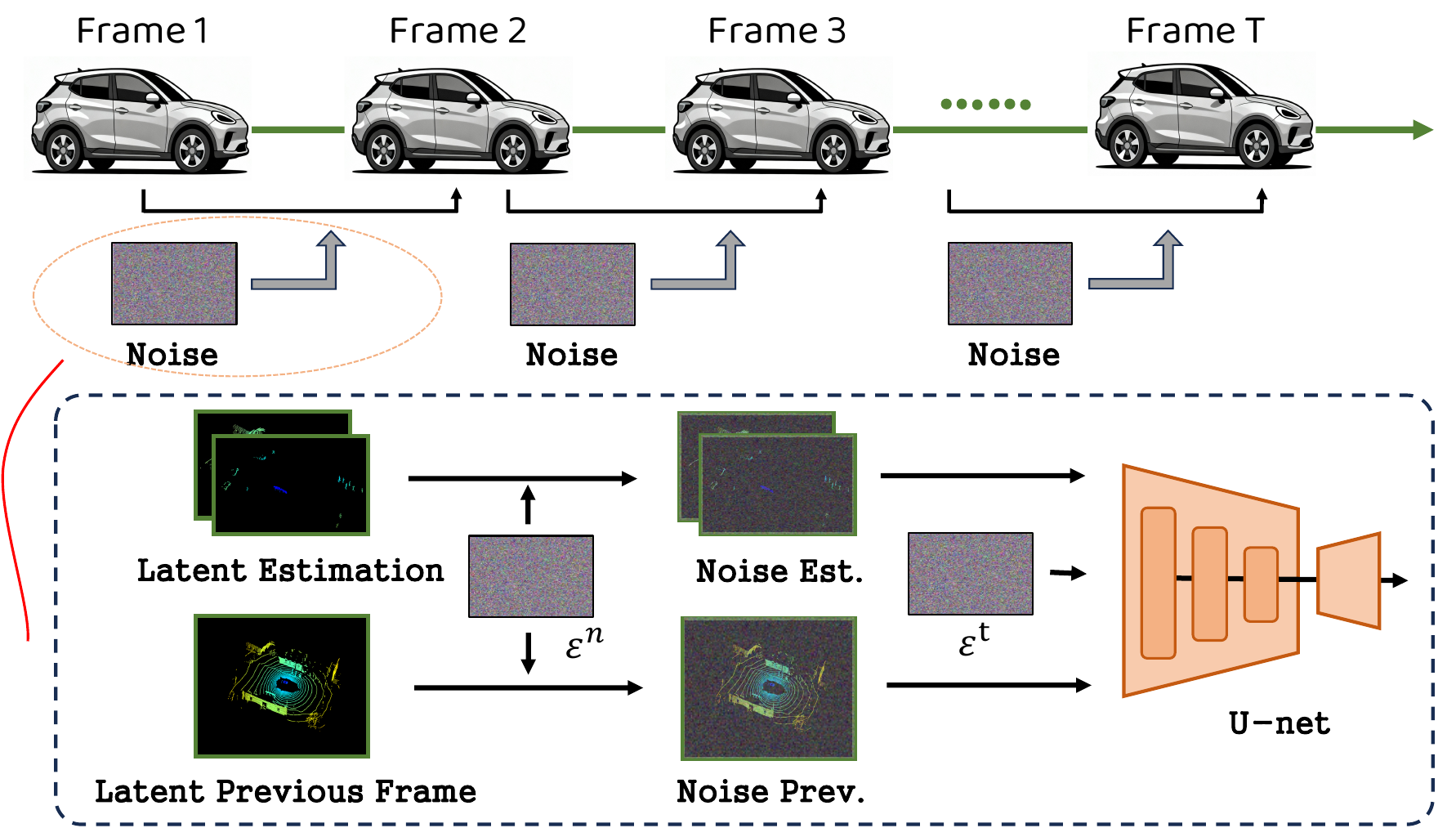}
    \caption{\label{nm_model}
    The schematic diagram of noise modulation module. }
\end{wrapfigure}
%------------------------------------------------------------------
As the recursive generation continues, these errors can accumulate, resulting in deviations from the actual situation when generating long-horizon data.

To this end, we design a module to mitigate error accumulation and enhance temporal consistency in long-horizon generation. 
The core idea of this module is to inject different levels of random noise into the previous frame’s image \cite{chen2024diffusion, deng2024streetscapes}, thereby reducing the model’s over-reliance on it.

As shown in Figure \ref{nm_model}, we add Gaussian noise with varying intensities to the latent feature \( \hat{z}^{s-1} \) corresponding to the previous frame’s LiDAR point cloud:
\begin{equation}
\label{diffusion_forcing}
\hat{z}_{\epsilon}^{s-1} =\sqrt{\overline{\alpha}_{n}} \cdot  \hat{z}^{s-1} +\sqrt{1-\overline{\alpha}_{n}} \cdot \epsilon,
\end{equation}
where \( \epsilon \sim \mathcal{N}(0, 1) \) is randomly sampled Gaussian noise, and \( n \in U[0, N] \) is the noise level.
Similarly, we inject noise to the encoded features of \( \widetilde{P}^{s}_{\text{obj}} \) and \( \widetilde{P}^{s}_{\text{bg}} \).

This strategy ensures better spatiotemporal consistency for the model when inferring long-horizon scenes, while only causing minor impact on short-horizon generation.
In the supplementary materials, we have provided a more detailed theoretical analysis of this module.

Upon integrating the two aforementioned modules into the overall architecture, we develop an inference framework for autoregressive LiDAR scene generation (see Figure~\ref{overview}).
In this process, the input LiDAR data is generated iteratively; in other words, the LiDAR data input at the current step is the output from the generator at the preceding step.
In the inference process, the output of the SDE module is also generated iteratively.
Consistent with the training phase, during inference the NM module also adds noise into the latent variables corresponding to these data.

%--------------------------------------------------------------
\section{Experiments}

\subsection{Experimental Setups}
\noindent\textbf{Dateset.}
We conducted experiments on the large-scale public nuScenes dataset \cite{caesar2020nuscenes}. 
It contains 1,000 autonomous driving sequences collected from the Boston and Singapore regions.
These two cities are known for their dense traffic and highly challenging driving conditions. 
The nuScenes training set contains 297,737 LiDAR scans, while the validation set contains 52,423 LiDAR scans, with each LiDAR scan comprising 32 beams.
This dataset is widely used for various tasks in autonomous driving, including but not limited to 3D object detection \cite{huang2021bevdet, li2023bevdepth, li2023end, li2024bevformer}, multi-object tracking \cite{hu2022monocular, pang2022simpletrack, zhang2022mutr3d}, trajectory prediction \cite{gu2023vip3d, liang2020pnpnet}, semantic occupancy prediction \cite{tong2023scene}, and open-loop planning for end-to-end autonomous driving \cite{hu2022st, hu2023planning, jiang2023vad}.
To verify the cross-dataset generalization ability of LaGen, we further conducted additional experiments on the denser-beam KITTI-360 dataset \cite{Geiger2013IJRR}.
Unlike the nuScenes dataset, each LiDAR scan in the KITTI-360 dataset contains 64 beams.

%--------------------------------------------------------------
\noindent\textbf{Baselines.}
We compare LaGen with three categories of baseline methods. 
The first category consists of traditional LiDAR data generation approaches, including LiDARGen \cite{LiDARGen2022}, LiDM \cite{ran2024towards}, and RangeLDM \cite{hu2024rangeldm}, with which we compare generation fidelity. 
The second category includes methods capable of generating temporally coherent scenes, such as UniScene \cite{li2025uniscene}, OpenDWM \cite{ni2025maskgwm}, and LiDARCrafter \cite{liang2026lidarcrafter}. 
In addition to comparing generation fidelity, we further evaluate temporal consistency metrics for each generated LiDAR frame. 
The third category comprises LiDAR prediction methods, including 4D-Occ \cite{khurana2023point} and ViDAR \cite{yang2024visual}, against which we conduct comprehensive comparisons on prediction-related metrics in long-horizon forecasting tasks.

%--------------------------------------------------------------
\noindent\textbf{Evaluation Metrics.}
We introduce multiple types of metrics to evaluate the fidelity and spatiotemporal consistency of the generated results.
For evaluating the performance of the generator, we follow the approach in \cite{LiDARGen2022} and use two metrics: Maximum Mean Discrepancy (MMD) and Jensen-Shannon Divergence (JSD).
Following LiDARCrafter \cite{liang2026lidarcrafter}, we employ TTCE and CTC as key metrics to assess the temporal consistency of our LiDAR sequence synthesis.
We also use the Chamfer Distance (CD) \cite{fan2017point} along with two additional error metrics based on ray depth, which quantify the difference between the generated point cloud for each frame and the corresponding ground truth in the driving scenario.

%--------------------------------------------------------------------
\noindent\textbf{Implementation Details.}
During training, our model is trained for 200 epochs with a total of 176,000 optimization steps, the learning rate is 1e-4.
We adopt the DDIM sampler \cite{song2020denoising} for denoising during training, with the total number of diffusion timesteps \( T \) set to 1024.
The UNet consists of three downsampling blocks (with the latter two containing Transformer layers) and three upsampling blocks (with the first two containing Transformer layers).
During the inference process, we perform 50 denoising steps to generate point clouds.

For more detailed descriptions of the metric computation, as well as the specific experimental configurations, please refer to the supplementary material.

%-------------------------------------------------------------------
\begin{figure}[t]
  \centering
  \includegraphics[width=\linewidth]{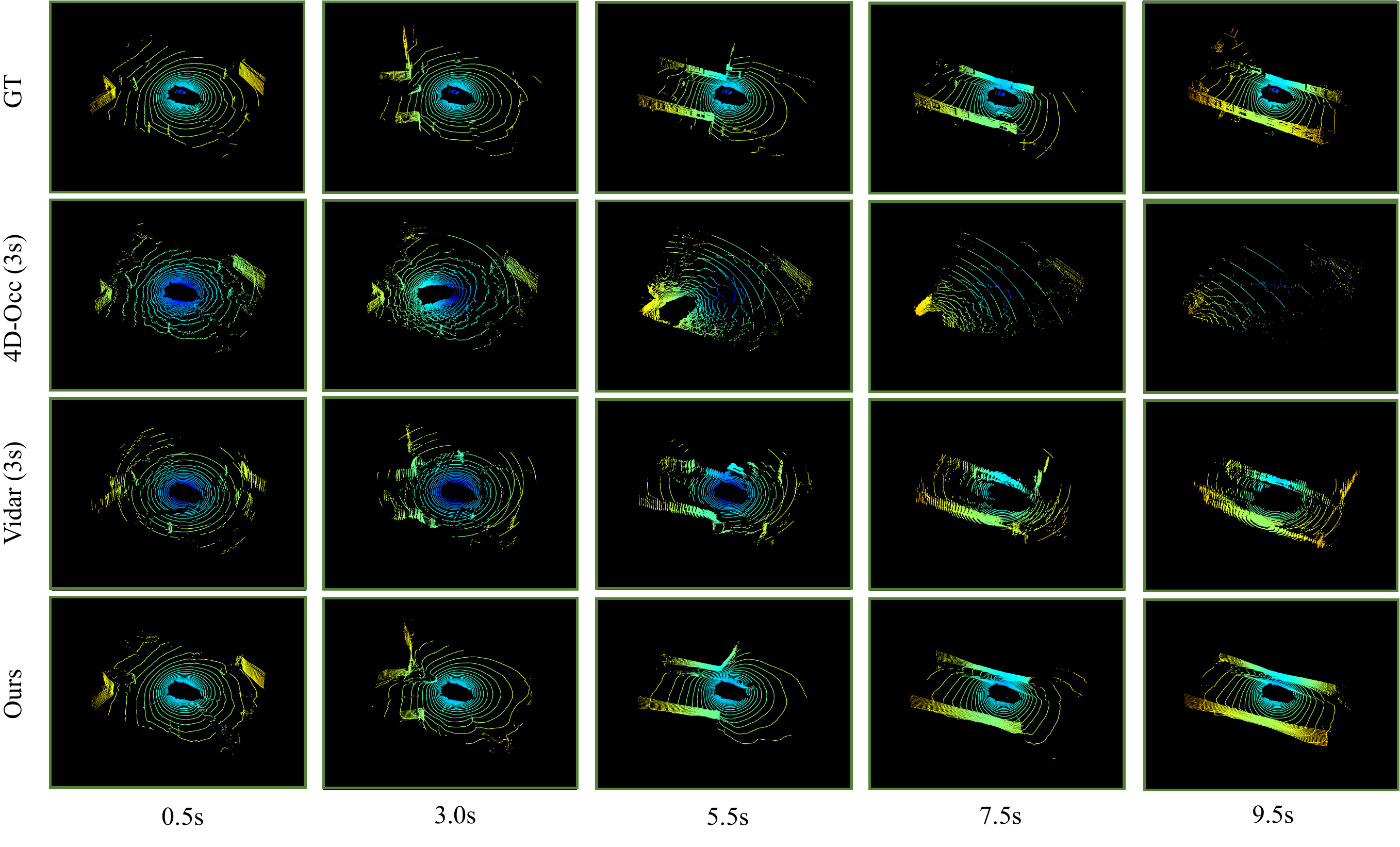}\\
  \caption{\label{results} 
    \textbf{Qualitative results.} Based on single-frame input, LaGen significantly outperforms other baseline methods in long-horizon prediction accuracy.}
\end{figure}
%-------------------------------------------------------------------

%--------------------------------------------------------------
\subsection{LiDAR Data Generation}
An important component of our approach is the LiDAR data generator, as the quality of its generated samples plays a crucial role in the overall framework performance.
Table \ref{table_generation} and Table \ref{table_generation_kitti} present the quantitative results of LaGen and several LiDAR generation baselines on the nuScenes and KITTI-360 datasets, respectively.
The experimental results show that LaGen outperforms these advanced baselines on several key metrics for LiDAR data generation.
In addition, using a subset of valid 3D bounding boxes from the nuScenes dataset, we further investigate the performance of LaGen on the downstream task and its object-level generation quality, as shown in Table \ref{table_downstream} and Table \ref{table_object_level}.
These results demonstrate that LaGen can effectively generate high-fidelity LiDAR data.

%--------------------------------------------------------------
\begin{table}[t]
\centering
\caption{The main quantitative results of LaGen and several baseline models on the nuScenes dataset. MMD values are reported in $10^{-4}$ and JSD in $10^{-2}$.}
\label{table_generation}
\setlength{\tabcolsep}{3.5pt}
\renewcommand{\arraystretch}{1.1}
\resizebox{0.5\columnwidth}{!}{
\small
\begin{tabular}{c|l|ccc} 
\toprule
& \textbf{Method} & \textbf{Venue} & \textbf{MMD $\downarrow$} & \textbf{JSD $\downarrow$} \\
\midrule \midrule 
\multirow{3}{*}{\rotatebox{90}{Uncond.}} 
& LiDARGen \cite{LiDARGen2022}   & ECCV'22 & 5.89 & 9.66  \\
& LiDM \cite{ran2024towards}       & CVPR'24 & 3.32 & 6.75  \\
& RangeLDM \cite{hu2024rangeldm}   & ECCV'24 & 1.92 & 5.47 \\
\midrule 
\multirow{4}{*}{\rotatebox{90}{Cond.}} 
& UniScene \cite{li2025uniscene}   & CVPR'25 & 2.40 & 10.80  \\
& OpenDWM \cite{ni2025maskgwm}    & CVPR'25 & 2.21 & 5.54  \\
& DriveLiDAR4D \cite{cai2026drivelidar4d} & AAAI'26 & 1.84 & 4.50  \\
& LiDARCrafter \cite{liang2026lidarcrafter} & AAAI'26 & 1.52 & 5.43 \\
& \cellcolor{myyellow} LaGen & \cellcolor{myyellow} -- & \cellcolor{myyellow} \textbf{0.11} & \cellcolor{myyellow} \textbf{2.77} \\
\bottomrule
\end{tabular}
}
\end{table}
%--------------------------------------------------------------

%--------------------------------------------------------------
\begin{table}[t]
\centering
\caption{The main quantitative results of LaGen and several baseline models on the KITTI-360 dataset. MMD values are reported in $10^{-4}$ and JSD in $10^{-2}$.}
\label{table_generation_kitti}
\setlength{\tabcolsep}{3.5pt}
\renewcommand{\arraystretch}{1.1}
\resizebox{0.60\linewidth}{!}{
\small
\begin{tabular}{l|ccccc} 
\toprule
\textbf{Method} 
& \textbf{Venue} 
& \textbf{MMD}$\downarrow$ 
& \textbf{JSD}$\downarrow$ 
& \textbf{FRD}$\downarrow$ 
& \textbf{FPD}$\downarrow$ \\
\midrule \midrule 
LiDARGen \cite{LiDARGen2022} & ECCV'22 & 13.36 & 13.67 & 2415.23 & 102.80 \\
LiDM \cite{ran2024towards}  & CVPR'24 & 19.00 & 9.31  & 2894.65 & 253.91 \\
OpenDWM \cite{ni2025maskgwm} & CVPR'25 & 14.63 & 12.93 & 1966.36 & 436.17 \\
\cellcolor{myyellow} LaGen    & \cellcolor{myyellow} --  & \cellcolor{myyellow} \textbf{1.41} & \cellcolor{myyellow} \textbf{5.75} 
 & \cellcolor{myyellow} \textbf{1109.83} & \cellcolor{myyellow} \textbf{73.50} \\
\bottomrule
\end{tabular}
}
\end{table}
%--------------------------------------------------------------

%--------------------------------------------------------------
\begin{table}[t]
\centering
\vspace{-2mm}

\begin{minipage}{0.35\columnwidth}
\centering
\caption{Results on the downstream 3D object detection task on the nuScenes dataset.}
\label{table_downstream}
\setlength{\tabcolsep}{5pt}
\renewcommand{\arraystretch}{1.1}
\resizebox{\linewidth}{!}{
\small
\begin{tabular}{l|cc} 
\toprule
\textbf{Method}  & \textbf{mAP}$\uparrow$ & \textbf{NDS}$\uparrow$\\
\midrule \midrule 
UniScene \cite{li2025uniscene}       & 0.51   & 0.47  \\
OpenDWM  \cite{ni2025maskgwm}       & 0.45   & 0.47  \\
LiDARCrafter  \cite{liang2026lidarcrafter}  & 0.41   & 0.41  \\
\cellcolor{myyellow} LaGen  & \cellcolor{myyellow} \textbf{0.52}  & \cellcolor{myyellow} \textbf{0.49} \\
\bottomrule
\end{tabular}
}
\end{minipage}
\hfill
\begin{minipage}{0.62\columnwidth}
\centering
\caption{Evaluation results on the quality of objects from all categories and selected categories in the generated LiDAR scenes on the nuScenes dataset.}
\label{table_object_level}
\setlength{\tabcolsep}{4pt}
\renewcommand{\arraystretch}{0.95}
\resizebox{\linewidth}{!}{
\small
\begin{tabular}{l|cc|cc|cc}
\toprule
\multirow{2}{*}{\textbf{Method}} 
& \multicolumn{2}{c|}{\textbf{Car}} 
& \multicolumn{2}{c|}{\textbf{Pedestrian}} 
& \multicolumn{2}{c}{\textbf{All Objects}} \\
\cmidrule(lr){2-3} \cmidrule(lr){4-5} \cmidrule(lr){6-7}
& \textbf{CD} $\downarrow$ & \textbf{EMD} $\downarrow$
& \textbf{CD} $\downarrow$ & \textbf{EMD} $\downarrow$
& \textbf{CD} $\downarrow$ & \textbf{EMD} $\downarrow$\\
\midrule
LiDM  \cite{ran2024towards}   & 3.563  & 3.763  & 0.701 & 0.663  & 3.236 & 4.241 \\
UniScene \cite{li2025uniscene} & 1.316  & 1.419  & 0.237 & 0.358  & 0.939 & 1.245 \\
LiDARCrafter \cite{liang2026lidarcrafter} & 3.657 & 2.581 & 0.636 & 0.557 & 3.347 & 3.517 \\
\cellcolor{myyellow} LaGen    
& \cellcolor{myyellow} \textbf{1.251} & \cellcolor{myyellow} \textbf{0.986} 
& \cellcolor{myyellow} \textbf{0.123} & \cellcolor{myyellow} \textbf{0.129} 
& \cellcolor{myyellow} \textbf{0.855} & \cellcolor{myyellow} \textbf{0.822} \\
\bottomrule
\end{tabular}
}
\end{minipage}
\end{table}
%--------------------------------------------------------------

%--------------------------------------------------------------
\subsection{Temporal LiDAR Scene Generation}
In this subsection, we evaluate the performance of LaGen in generating sequential LiDAR scenes. 
Following the metrics established in LiDARCrafter \cite{liang2026lidarcrafter}, we conduct a comprehensive assessment of various methods across the dimension of temporal consistency, as summarized in Table~\ref{metric_temporal_consistency}. 
The results demonstrate that LaGen achieves a significant improvement in the CTC metric while maintaining a substantial advantage in terms of TTCE.
In particular, LaGen can not only effectively generate long-sequence LiDAR scenes, but also maintain strong performance over extended generation horizons.

%--------------------------------------------------------------
\begin{table}[t]
\caption{The quantitative results on temporal consistency for 4D LiDAR generation, primarily evaluated using the TTCE and CTC metrics.}
\label{metric_temporal_consistency}
\setlength{\tabcolsep}{3.5pt}
\renewcommand{\arraystretch}{1.1}
\centering
\resizebox{0.7\columnwidth}{!}{
\small 
\begin{tabular}{@{}l|c|cc|ccccc@{}}
\toprule
\multirow{2}{*}{\textbf{Method}} & \multirow{2}{*}{\textbf{Venue}} & \multicolumn{2}{c|}{\textbf{TTCE}$\downarrow$} & \multicolumn{5}{c}{\textbf{CTC}$\downarrow$} \\
 &  & \textbf{3} & \textbf{4} & \textbf{1} & \textbf{2} & \textbf{3} & \textbf{4} & \textbf{10}\\ \midrule \midrule
UniScene \cite{li2025uniscene} & CVPR'25 & 2.74 & 3.69 & 0.90 & 1.84 & 3.64 & \cellcolor{Lavender} 3.90 & -\\
OpenDWM \cite{ni2025maskgwm} & CVPR'25 & \cellcolor{Lavender} 2.68 & 3.65 & 1.02 & 2.02 & 3.37 & 5.05 & -\\
OpenDWM-DiT \cite{ni2025maskgwm} & CVPR'25 & 2.71 & 3.66 & \cellcolor{Lavender} 0.89 & \cellcolor{Lavender} 1.79 & 3.06 & 4.64 & -\\ 
LiDARCrafter \cite{liang2026lidarcrafter} & AAAI'26 & \cellcolor{myred} \textbf{2.65} & \cellcolor{myred} \textbf{3.56} & 1.12 & 2.38 & \cellcolor{Lavender} 3.02 & 4.81 & - \\ \midrule
\textbf{LaGen} & \textbf{--} & 2.71 & \cellcolor{Lavender} \textbf{3.62} & \cellcolor{myred} \textbf{0.60} & \cellcolor{myred} \textbf{0.97} & \cellcolor{myred} \textbf{1.46} & \cellcolor{myred} \textbf{2.22} & \cellcolor{myred} \textbf{3.84} \\ \bottomrule
\end{tabular}
}
\end{table}
%--------------------------------------------------------------

%--------------------------------------------------------------
\subsection{Long-Horizon Prediction}
To further validate the long-horizon autoregressive performance of LaGen on other tasks, we designed two distinct modes and conducted experiments on predictive tasks.
Specifically, we evaluate two variants: 
1) a box-free LaGen baseline to assess the model’s mastery of underlying scene dynamics, 
and 2) the full LaGen framework to demonstrate how integrating current feedback bolsters accuracy in long-term autoregressive forecasting.
We have provided the results of several state-of-the-art predictive models for reference, as shown in Table \ref{table_prediction}.
These results validate that LaGen yields outstanding performance under a predictive paradigm more conducive to practical deployment (see Figure~\ref{results}).

%--------------------------------------------------------------
\setlength{\tabcolsep}{4pt} 
\begin{table*}[t]
\centering
\caption{Comparison of LaGen with several state-of-the-art LiDAR prediction models. The underlined data represents the results obtained by introducing rolling inference on the original 4D-Occ model. ViDAR requires visual inputs during prediction.}
\renewcommand{\arraystretch}{1.5}
\resizebox{\textwidth}{!}{ 
\begin{tabular}{cc|cccccc|cccccc|cccccc}
\toprule
\multirow{2}{*}{\textbf{Method}} & \textbf{Input} & \multicolumn{6}{c|}{\textbf{Chamfer Distance (m$^2$)$\downarrow$}} & \multicolumn{6}{c|}{\textbf{L1 Error (m) $\downarrow$}} & \multicolumn{6}{c}{\textbf{Absrel (\%)$\downarrow$}} \\
\cmidrule(lr){3-8} \cmidrule(lr){9-14} \cmidrule(lr){15-20}
& \textbf{frames} & 0.5s & 1.5s & 3.5s & 5.5s & 7.5s & 9.5s & 0.5s & 1.5s & 3.5s & 5.5s & 7.5s & 9.5s & 0.5s & 1.5s & 3.5s & 5.5s & 7.5s & 9.5s \\
\midrule
4D-Occ \cite{khurana2023point}
& 2 
& 1.15 & \underline{1.84} & \underline{4.30} & \underline{12.30} & \underline{25.11} &\underline{55.47}  
& 1.24 & \underline{2.04} & \underline{3.41} & \underline{4.87} & \underline{7.07} &\underline{9.95}
& 8.88 & \underline{14.82} & \underline{25.36} & \underline{40.14} & \underline{56.24} & \underline{67.37}\\
4D-Occ \cite{khurana2023point}
& 6 
& \cellcolor{Lavender} 1.10 & \cellcolor{Lavender} 1.33 &  \underline{2.41} & \underline{7.73} & \underline{12.82} & \underline{21.08} 
& \cellcolor{Lavender} 1.23 & \cellcolor{Lavender} 1.62 & \cellcolor{Lavender} \underline{2.60} & \cellcolor{Lavender} \underline{3.48} & \underline{4.74} & \underline{6.49}
& \cellcolor{Lavender} 8.41 & \cellcolor{Lavender} 12.15 & \cellcolor{Lavender} \underline{22.16} & \cellcolor{Lavender} \underline{29.20} & \underline{35.78} & \underline{39.50} \\
\midrule
ViDAR \cite{yang2024visual} 
& 2 
& 1.15 & 1.56 & 2.50 & 3.79 & 5.18 & 6.37
& 2.38 & 2.70 & 3.34 & 4.24 & 5.40 & 6.06 
& 16.29 & 19.96 & 27.34 & 36.25 & 48.61 & 54.13 \\
ViDAR \cite{yang2024visual} 
& 6
& 1.14 & 1.47 & \cellcolor{Lavender} 2.30 & \cellcolor{Lavender} 3.25 & \cellcolor{Lavender} 4.12 & \cellcolor{Lavender} 5.06
& 2.15 & 2.60 & 3.06 & 3.49 & \cellcolor{Lavender} 3.91 & \cellcolor{Lavender} 4.42 
& 17.98 & 20.57 & 25.87 & 30.18 & \cellcolor{Lavender}  33.77 & \cellcolor{Lavender}  38.34 \\
\midrule
\textbf{Ours(w/o BBox)} & 
\cellcolor{gray!20}1
& \cellcolor{myred} \textbf{0.65} & \cellcolor{myred} \textbf{0.93} & \cellcolor{myred} \textbf{1.44} & \cellcolor{myred} \textbf{1.98} & \cellcolor{myred} \textbf{2.39} & \cellcolor{myred} \textbf{2.78}  
& \cellcolor{myred} \textbf{0.14} & \cellcolor{myred} \textbf{0.20} & \cellcolor{myred} \textbf{0.26} & \cellcolor{myred} \textbf{0.31} & \cellcolor{myred} \textbf{0.36} & \cellcolor{myred} \textbf{0.40}
& \cellcolor{myred} \textbf{2.15} & \cellcolor{myred} \textbf{2.66} &  \cellcolor{myred} \textbf{3.23} & \cellcolor{myred} \textbf{3.73} & \cellcolor{myred} \textbf{4.22} & \cellcolor{myred} \textbf{4.42}\\
\midrule
\textbf{Ours(w/ BBox)} & 
\cellcolor{gray!20}1
& \cellcolor{myred} \textbf{0.50} & \cellcolor{myred} \textbf{0.72} & \cellcolor{myred} \textbf{1.25} & \cellcolor{myred} \textbf{1.69} & \cellcolor{myred} \textbf{2.09} & \cellcolor{myred} \textbf{2.34}  
& \cellcolor{myred} \textbf{0.13} & \cellcolor{myred} \textbf{0.18} & \cellcolor{myred} \textbf{0.24} & \cellcolor{myred} \textbf{0.28} & \cellcolor{myred} \textbf{0.30} & \cellcolor{myred} \textbf{0.32}
& \cellcolor{myred} \textbf{2.04} & \cellcolor{myred} \textbf{2.47} &  \cellcolor{myred} \textbf{3.03} & \cellcolor{myred} \textbf{3.40} & \cellcolor{myred} \textbf{3.63} & \cellcolor{myred} \textbf{3.61}\\
\bottomrule
\end{tabular}
}
\label{table_prediction}
\end{table*}
%--------------------------------------------------------------

%--------------------------------------------------------------
\subsection{Editability and Interactive Generation}

%------------------------------------------------------------------
\begin{wrapfigure}[17]{r}{0.51\textwidth}
    \centering
    \includegraphics[width=0.95\linewidth]{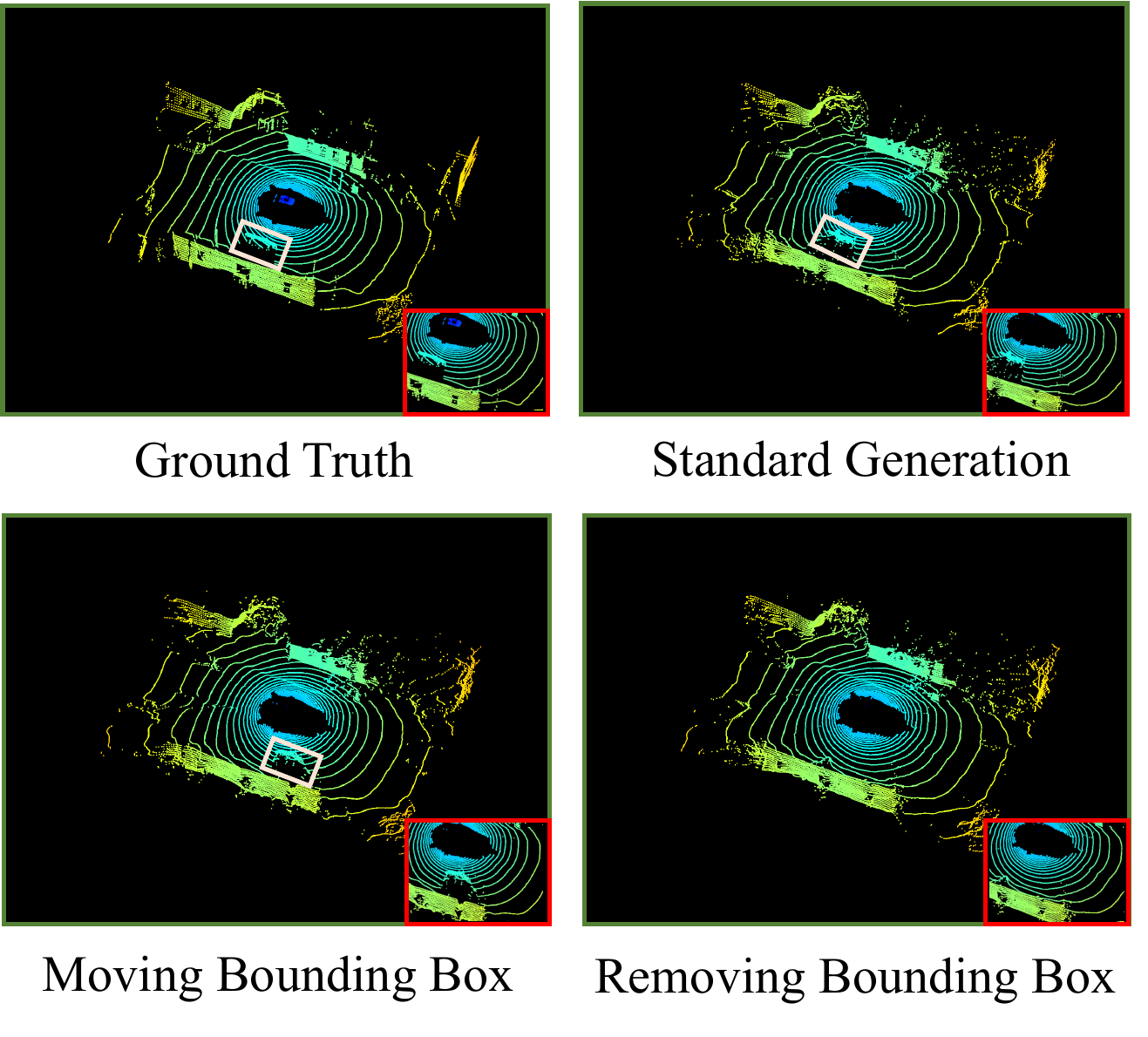}
    \caption{\label{interaction} Visualization of object-level editing. }
\end{wrapfigure}
%------------------------------------------------------------------

%--------------------------------------------------------------
\begin{table}[t]
\centering

\begin{minipage}[t]{0.49\textwidth}
\renewcommand{\arraystretch}{1.5}
\centering
\caption{The ablation results on the scene decoupling estimation module of LaGen.}
\label{table_ablation_sde}
\resizebox{\linewidth}{!}{
\small
\begin{tabular}{@{}l|cccc|cccccc@{}}
\toprule
\multirow{2}{*}{\textbf{LaGen}} & \multicolumn{4}{c|}{\textbf{CTC}$\downarrow$} & \multicolumn{6}{c}{\textbf{Chamfer Distance (m$^2$)}$\downarrow$} \\
\cmidrule(lr){2-5} \cmidrule(l){6-11}
 & \textbf{1} & \textbf{2} & \textbf{3} & \textbf{4} 
 & \textbf{0.5s} & \textbf{1.0s} & \textbf{2.0s} & \textbf{4.0s} & \textbf{6.0s} & \textbf{8.0s} \\
\midrule \midrule
w/o SDE  & 0.71 & 1.15 & 1.66 & 2.49 & 0.56 & 0.66 & 0.95 & 1.42 & 1.87 & 2.23 \\
\rowcolor{myyellow}
w/ SDE   & \textbf{0.60} & \textbf{0.97} & \textbf{1.46} &  \textbf{2.22} & \textbf{0.50} & \textbf{0.61} & \textbf{0.83} & \textbf{1.36} & \textbf{1.80} & \textbf{2.16}  \\
\bottomrule
\end{tabular}
}
\end{minipage}
\hfill
\begin{minipage}[t]{0.49\textwidth}
\renewcommand{\arraystretch}{1.5}
\centering
\caption{The ablation results on the noise modulation module of LaGen.}
\label{table_ablation_nm}
\resizebox{\linewidth}{!}{
\small
\begin{tabular}{@{}l|cccc|cccccc@{}}
\toprule
\multirow{2}{*}{\textbf{LaGen}} & \multicolumn{4}{c|}{\textbf{CTC}$\downarrow$} & \multicolumn{6}{c}{\textbf{Chamfer Distance (m$^2$)}$\downarrow$} \\
\cmidrule(lr){2-5} \cmidrule(l){6-11}
 & \textbf{1} & \textbf{2} & \textbf{3} & \textbf{4} 
 & \textbf{4.0s} & \textbf{5.0s} & \textbf{6.0s} & \textbf{7.0s} & \textbf{8.0s} & \textbf{9.0s} \\
\midrule \midrule
w/o NM  & 0.73 & 1.21 & 1.73 & 2.77 & 1.38 & 1.79 & 2.09 & 2.50 & 2.82 & 3.16 \\
\rowcolor{myyellow}
w/ NM  & \textbf{0.60} & \textbf{0.97} & \textbf{1.46} &  \textbf{2.22} & \textbf{1.36} & \textbf{1.66} & \textbf{1.80} & \textbf{2.02} & \textbf{2.16} & \textbf{2.25} \\
\bottomrule
\end{tabular}
}
\end{minipage}
\end{table}
%--------------------------------------------------------------

In this subsection, we demonstrate LaGen’s interactive generation capabilities. 
Since it generates scenes frame by frame, we can edit the positions of objects in the scene at any intermediate frame.
In Figure~\ref{interaction}, we illustrate interactive generation by moving or removing a bounding box corresponding to a vehicle.
We observe that after moving the other vehicle, LaGen accurately restore the occlusion effects caused by the object's new position. 
Additionally, when the vehicle is removed, LaGen successfully completes the previously occluded portions of the road during standard generation.
The interactivity of LaGen in autoregressive generation represents a significant advancement, enabling it to generate more realistic 4D scenes according to changes in the external environment.
Moreover, this framework naturally supports diverse LiDAR scene generation, as different possible outcomes can be achieved via stochastic or user-defined editing.

%--------------------------------------------------------------
\subsection{Ablation Study}

We perform ablation studies to evaluate the contributions of the scene decoupling estimation and the noise
modulation modules. 
As shown in Table \ref{table_ablation_sde}, the SDE module effectively enhances temporal consistency between consecutive frames during scene generation. In addition, it improves the accuracy of long-horizon autoregressive tasks to a certain extent.
Table \ref{table_ablation_nm} demonstrates that the NM module enhances temporal consistency and markedly alleviates error accumulation in long-horizon LiDAR scene generation, with increasingly significant improvements over extended time horizons.

We also investigate the impact of different denoising steps during inference on both accuracy and latency (see Table~\ref{table_ablation_denoising}). 
As the number of denoising steps increases, the performance on long-horizon generation tasks improves to some degree, at the cost of increased average inference time per frame. Notably, the marginal performance gains decrease as the number of denoising steps becomes larger.
When near-real-time inference performance is required, we can choose a smaller number of denoising steps. 
As shown in Table~\ref{table_ablation_denoising}, using fewer denoising steps can substantially improve inference speed while only slightly affecting model performance.
Finally, we conducted ablation experiments at the inference stage by randomly perturbing and dropping the bounding box information to evaluate the robustness of the model; see the supplementary material for details.

%--------------------------------------------------------------
\begin{table}[t]
\centering
\caption{The ablation results of LaGen on the number of denoising steps and the corresponding time consumption during inference.}
\setlength{\tabcolsep}{3.5pt}
\renewcommand{\arraystretch}{1.1}
\resizebox{0.6\columnwidth}{!}{
\small
\begin{tabular}{cccccccc}
\toprule
\textbf{Denoising} &
\multicolumn{6}{c}{\textbf{Chamfer Distance (m$^2$)}} &
\textbf{Time} \\
\cmidrule(l{3pt}r{3pt}){2-7}

\textbf{Steps} &
0.5s & 1.0s & 3.0s & 5.0s & 7.0s & 9.5s &
(s/frame) \\
\midrule
10
& 0.64 & 0.79 & 1.44 & 2.12 & 2.35 & 2.60
& 0.21 \\
%\midrule
30
& 0.52 & 0.64 & 1.16 & 1.83 & 2.09 & 2.41
& 0.33 \\
%\midrule
\rowcolor{myyellow}
50
& 0.50 & 0.61 & 1.11 & 1.66 & 2.02 & 2.34
& 0.43 \\
%\midrule
100
& 0.50 & 0.59 & 1.06 & 1.72 & 2.12 & 2.31
& 0.73 \\
%\midrule
200
& 0.50 & 0.59 & 1.03 & 1.51 & 2.10 & 2.20
& 1.44 \\
\bottomrule

\end{tabular}
\label{table_ablation_denoising}
}
\vspace{-2mm}
\end{table}
%--------------------------------------------------------------

%--------------------------------------------------------------
\subsection{Limitation and Discussion}
LaGen still has the following limitations.
First, LaGen mainly targets common single-return spinning mechanical LiDAR and does not explicitly model multi-return LiDAR sensors.
Since multiple returns still lie along the same ray, a possible future extension to this setting would be to assign a separate depth channel for each return in the range-view representation. 
For non-spinning LiDAR, new representations of LiDAR scenes need to be further explored.
Second, in the SDE module, to ensure consistency across different task settings, we adopt a nearest-distance-based search mechanism.
However, in crowded scenes with multiple objects of the same category, this may affect the effectiveness of the module to some extent. 
In the future, tracklets could be used in simulation tasks to match object boxes across adjacent frames.

%--------------------------------------------------------------

%--------------------------------------------------------------
\section{Conclusion}

We propose LaGen, a novel framework capable of autoregressively generating long-horizon LiDAR scenes in a frame-by-frame manner.
We develop a LiDAR data generator conditioned on multiple control conditions within the driving scenario.
We further enhance the spatiotemporal consistency of generation through the scene decoupling estimation module and the noise modulation module.
We thoroughly evaluate the performance of LaGen on the nuScenes and KITTI-360 datasets. 
The experimental results demonstrate the effectiveness of this framework and its advantages in long-horizon generation tasks.
In the future, we expect LaGen to play an important role in applications such as closed-loop simulation and world modeling, thereby advancing the development of autonomous driving.

\section*{Acknowledgments}
This research was supported by the Scientific Research Innovation Capability Support Project for Young Faculty (U40) of the Ministry of Education of China (SRICSPYF-ZY2025019), the National Natural Science Foundation of China (No.62272433), Anhui Provincial Natural Science Foundation (No.2508085ZD011), Key Science \& Technology Project of Anhui Province (No. 202523o09050004) and the Fundamental Research Funds for the Central Universities.

\clearpage  % TODO FINAL: This \clearpage needs to be removed from both review and camera-ready versions.

\bibliographystyle{splncs04}
\bibliography{main}

%\iffalse
%%%%%%%%%%%%%%%%%%%%%%%%%%%%%%%%%%%%%%%%%%%%%%%%%%%%%%%%%%%%%%%%%%%%%%%%%%%%%%%
%%%%%%%%%%%%%%%%%%%%%%%%%%%%%%%%%%%%%%%%%%%%%%%%%%%%%%%%%%%%%%%%%%%%%%%%%%%%%%%
% APPENDIX
%%%%%%%%%%%%%%%%%%%%%%%%%%%%%%%%%%%%%%%%%%%%%%%%%%%%%%%%%%%%%%%%%%%%%%%%%%%%%%%
%%%%%%%%%%%%%%%%%%%%%%%%%%%%%%%%%%%%%%%%%%%%%%%%%%%%%%%%%%%%%%%%%%%%%%%%%%%%%%%
\newpage
\appendix
\onecolumn
\newcommand{\respond}[3]{
	\begin{description}[leftmargin=2em,itemsep=0pt,parsep=4pt,topsep=12pt]
		\item[Q{#1}:] \emph{#2}
		\item[A{#1}:] {#3}
	\end{description}
}

\title{Supplementary Material} 

\titlerunning{LaGen}

\author{}
\institute{}

\authorrunning{S. Zhou et al.}

\maketitle

%--------------------------------------------------------------------
\section{Overview}
In the appendix, we provide supplementary content for the main text organized into the following sections:
\begin{itemize}[leftmargin=*]
\item More implementation details;
\item Detailed explanations of the evaluation metrics, along with their formal definitions and formulas;
\item A thorough introduction to the principles of range-view projection;
\item Detailed theoretical analysis of the noise modulation module;
\item Detailed theoretical analysis of the scene decoupling estimation module;
\item Visualization of SDE module inputs and outputs;
\item Ablation results on randomly perturbing or dropping the 3D bounding box;
\item Additional quantitative results, presented frame by frame;
\item Additional qualitative results, provided on a per-frame basis, showcasing two representative and substantially different scenarios.

\end{itemize}

%--------------------------------------------------------------------
\section{More Implementation Details}
During training, our model is trained for 200 epochs with a total of 176,000 optimization steps. 
The overall batch size is set to 32, the learning rate is 1e-4, and a cosine annealing scheduler is employed.
The model is trained in a distributed manner on 4 NVIDIA H200 GPUs, requiring approximately 23 hours in total.
We adopt the DDIM sampler \cite{song2020denoising} for denoising during training, with the total number of diffusion timesteps \( T \) set to 1024.
The VAE in LDM adopts pretrained weights from RangeLDM \cite{hu2024rangeldm}. 
The range image has a resolution of [1024,32], and the VAE employs a downsampling factor of 4.
The UNet consists of three downsampling blocks (with the latter two containing Transformer layers) and three upsampling blocks (with the first two containing Transformer layers).
To better handle the periodic structure of range-view representations, all standard convolutional layers in both the VAE and UNet are replaced with circular convolutions.
During inference, we use the DDIM sampler to perform 50 denoising steps for point cloud generation.
Inference was performed on a single NVIDIA H200 GPU.

%--------------------------------------------------------------
\section{More Details of Evaluation Metrics}
In the main text, we introduce Maximum Mean Discrepancy (MMD) and Jensen-Shannon Divergence (JSD) to evaluate the performance of the generator. 
Below, we provide a detailed explanation of these two metrics.

Maximum Mean Discrepancy (MMD) is a non-parametric method for measuring the distance between two probability distributions. 
The fundamental idea is to compute the distance between the kernel means of two sample sets in a high-dimensional space; by comparing the means mapped to this space, one can determine whether the sample distributions are consistent:
\begin{equation}
\label{mmd}
    MMD = \frac{1}{N^2} \sum_{i}^{N} \sum_{j}^{N} k(p_i, p_j) + \frac{1}{N^2} \sum_{i}^{N} \sum_{j}^{N} k(q_i, q_j) - \frac{2}{N^2} \sum_{i}^{N} \sum_{j}^{N} k(p_i, q_j).
\end{equation}

Jensen-Shannon Divergence (JSD) is a symmetric information-theoretic measure for quantifying the similarity between two probability distributions, derived from the Kullback-Leibler (KL) divergence and commonly used for distribution comparison. 
The essence of JSD is to evaluate the information loss of each distribution with respect to their uniform mixture:
\begin{equation}
\label{jsd}
JSD = \frac{1}{2}\sum_{i}^{N} (p_i \log\frac{p_i}{m_i} + q_i \log\frac{q_i}{m_i}).
\end{equation}
We compute the MMD and JSD metrics using 100×100 2D histograms on the bird’s-eye view (BEV) plane.

To evaluate the temporal consistency of the synthesized sequences, we follow LiDARCrafter~\cite{liang2026lidarcrafter} and adopt the TTCE (Translation and Transformation Consistency Error) and CTC (Chamfer Temporal Consistency) metrics.

Let $T_{t}^{gt}=[R_{t}^{gt} \mid t_{t}^{gt}]$ denote the ground-truth transformation from frame $t$ to $t+1$, where $R_{t}^{gt}$ represents the rotation matrix and $t_{t}^{gt}$ is the translation vector. 
Let $T_{t}^{pred}=[R_{t}^{pred} \mid t_{t}^{pred}]$ be the transformation estimated from the generated point clouds via Iterative Closest Point (ICP). 
The TTCE is calculated as follows:
\begin{equation}
TTCE = \frac{1}{T-1} \sum_{t=1}^{T-1} \left| t_{t}^{pred} - t_{t}^{gt} \right|_2, 
\end{equation}
where $\|\cdot\|_F$ denotes the Frobenius norm and $T$ is the total number of frames in the sequence.

Furthermore, let $\mathcal{P}_t$ and $\mathcal{P}_{t+1}$ be the generated point clouds at frames $t$ and $t+1$, respectively. Using the ground-truth transformation $T_{t}^{gt}$, we align $\mathcal{P}_{t+1}$ to the coordinate system of frame $t$:
\begin{equation}
\tilde{\mathcal{P}}_{t+1} = \left(R_{t}^{gt}\right)^{-1} \left( \mathcal{P}_{t+1} - t_{t}^{gt} \right). 
\end{equation}
Subsequently, the Chamfer Distance (CD) between $\mathcal{P}_t$ and the aligned $\tilde{\mathcal{P}}_{t+1}$ is computed:
\begin{equation}
 CD\left(\mathcal{P}_t, \tilde{\mathcal{P}}_{t+1}\right) = \frac{1}{\left|\mathcal{P}_t\right|} \sum_{x \in \mathcal{P}_t} \min_{y \in \tilde{\mathcal{P}}_{t+1}} \| x - y \|_2^2 + \frac{1}{\left|\tilde{\mathcal{P}}_{t+1}\right|} \sum_{y \in \tilde{\mathcal{P}}_{t+1}} \min_{x \in \mathcal{P}_t} \| y - x \|_2^2. 
\end{equation}
The final CTC score is defined as the average Chamfer Distance across all adjacent frame pairs:
\begin{equation}
CTC = \frac{1}{T-1} \sum_{t=1}^{T-1} CD\left(\mathcal{P}_t, \tilde{\mathcal{P}}_{t+1}\right).
\end{equation}

In addition, we introduce Chamfer Distance (CD) \cite{fan2017point} and two error metrics based on ray depth to evaluate the accuracy of autoregressive LiDAR generation.
Chamfer Distance is a commonly used metric for measuring similarity between point clouds. 
It quantifies the “alignment” distance between two sets of points, representing the sum of the distances from each point in one set to its nearest neighbor in the other set. 
By aggregating these nearest neighbor distances across all points, Chamfer Distance captures the overall distribution difference:
\begin{equation}
\label{Chamfer_Distance}
    CD(P^{s}_{\text{gen}}, P^{s}_{\text{gt}})= \frac{1}{M} \sum_{x \in P^{s}_{\text{gt}}} \min_{y \in P^{s}_{\text{gen}}} ||x - y||_2^2 + \frac{1}{N} \sum_{y \in P^{s}_{\text{gen}}} \min_{x \in P^{s}_{\text{gt}}} ||y - x||_2^2.
\end{equation}

The L1 metric measures the mean absolute error between the predicted depth and the ground truth depth along each ray, directly reflecting the magnitude of absolute deviation in depth prediction.
It is calculated by averaging the absolute error between the rendered depth \( \lambda' \) and the ground truth depth \( \lambda \) for all query rays:
\begin{equation}
L1\_Error=\frac{1}{N} \sum_{i=1}^{N} |\lambda_i - \lambda'_i|
\end{equation}

The AbsRel metric (Absolute Relative Error) reflects the degree of relative deviation in depth prediction and is particularly suitable for evaluating the accuracy of near-field depth estimation:
\begin{equation}
AbsRel=\frac{1}{N} \sum_{i=1}^{N} \frac{|\lambda_i - \lambda'_i|}{\lambda_i}
\end{equation}
where \( N \) is the total number of query rays, \( \lambda_i \) and \( \lambda_i' \) denote the ground truth depth and the predicted depth for the \( i \)-th ray, respectively.

Considering the actual requirements of autonomous driving scenarios, we measured the above indicators within the range of [-50.0, 50.0] m.

%------------------------------------------------------------------
\section{Comparison between standard range-view projection and the corrected version}
Range images offer a compact representation of LiDAR data by parametrizing unstructured point clouds in 3D space into a 2D pixel space. 
This projection process is reversible, and after applying it, both training and inference costs for the model can be significantly reduced. Consequently, representing LiDAR data with range images has become one of the prevailing trends in the field.

The standard range image mapping method projects each point \( p = (x, y, z) \in \mathbb{R}^3 \) in the point cloud into the spherical coordinates \( (r, \theta, \phi) \) of the range image using spherical coordinate projection:
\begin{equation}
\label{range_view}
\begin{aligned}
    r &= \sqrt{x^{2} + y^{2} + z^{2}}, \\
    \theta &= \arctan(y, x), \\
    \phi &= \arctan\left(z, \sqrt{x^{2} + y^{2} + z^{2}}\right),
\end{aligned}
\end{equation}
where \( r \) is the depth value, \( \theta \) is the azimuth angle, and \( \phi \) is the elevation angle.
RangeLDM \cite{hu2024rangeldm} points out that multiple lasers in a LiDAR system do not share the same origin during measurement, which can introduce errors during coordinate transformation and lead to incorrect range-view data distributions. 
Therefore, we adopt its corrected range image representation as follows:
\begin{equation}
\label{corrected_range_view}
\begin{aligned}
    r &= \sqrt{x^{2} + y^{2} + (z - h_{j})^{2}}, \\
    \theta &= \arctan(y, x), \\
    \phi &= \phi_{j},
\end{aligned}
\end{equation}
where \( h_{j} \) and \( \phi_{j} \) represent the sensor height and elevation angle estimated by Hough voting, respectively.
In subsequent computations, both projection methods are processed in exactly the same way, including converting spherical coordinates \( (r, \theta, \phi) \) to range image pixel coordinates \( (u, v) \):
\begin{equation}
\begin{aligned}
    u &= W - ((\theta + \pi) / 2\pi) \cdot W, \\
    v &= H - j,
\end{aligned}
\end{equation}
as well as normalizing the depth and intensity values of the point cloud to obtain the two-channel range image \( I \).

In the subsequent calculation of the CD metric, it is necessary to reverse-convert the range image back into a point cloud. 
In the corrected version, the standard LiDAR's fifth dimension, which is the line index \( j \) for each point, is directly used during the forward projection process.
However, the generated point cloud contains only three-dimensional coordinates and intensity information. 
Therefore, we introduce the following formula to compute the line index for each point:
\begin{equation}
\begin{aligned}
    j = (\text{Rad2Deg}(\arcsin(\frac{z}{d})) - fov_{\text{down}}) / (fov_{\text{up}} - fov_{\text{down}}) \cdot (H - 1),
\end{aligned}
\end{equation}
where \( fov_{\text{up}} \) and \( fov_{\text{down}} \) refer to the elevation angle range of the original LiDAR point cloud acquisition.

%------------------------------------------------------------------
\section{Detailed theoretical analysis of the noise modulation module}

\subsection{Recursive Dynamics Model of Error Accumulation}
In standard autoregressive generation, the optimization objective of the model is to minimize the following:
\begin{equation}
\mathcal{L}_{clean} = \mathbb{E} [\| z_t - f(z_{t-1} ; \theta) \|^2].
\end{equation}
However, during the inference phase, the input is the predicted value contaminated by the preceding error $\delta_{t-1} = \hat{z}_{t-1} - z_{t-1}$. Through a first-order Taylor expansion, the error at frame $t$ can be approximated as:
\begin{equation}
\delta_t = \hat{z}_t - z_t = f(z_{t-1} + \delta_{t-1}) - f(z_{t-1}) \approx \mathbf{J}_f(z_{t-1}) \cdot \delta_{t-1},
\end{equation}
where $\mathbf{J}_f$ denotes the Jacobian Matrix of the function $f$.
If the maximum eigenvalue (spectral radius) of the Jacobian matrix satisfies $\rho(\mathbf{J}_f) > 1$, the error $\delta_t$ will exhibit exponential growth over time $t$. 
Given the high dimensionality and intricate local structures inherent in LiDAR data, an excessive spectral radius of $\mathbf{J}_f$ will inevitably introduce instability into the model.

\subsection{Argument for Expected Contraction under Noise Modulatioon}
When noise $\epsilon \sim \mathcal{N}(0, \mathbf{I})$ is introduced and the model is trained using $z'_{t-1} = \sqrt{\bar{\alpha}_n} z_{t-1} + \sqrt{1-\bar{\alpha}_n} \epsilon$, the loss function becomes:
\begin{equation}
\mathcal{L}_{noise} = \mathbb{E}_{z, \epsilon, n} \left[ \| z_t - f(\sqrt{\bar{\alpha}_n} z_{t-1} + \sqrt{1-\bar{\alpha}_n} \epsilon ; \theta) \|^2 \right]
\end{equation}
By performing a second-order Taylor expansion of the above loss function with respect to the noise intensity $\sigma_n^2 = 1-\bar{\alpha}_n$ (assuming small noise):
\begin{equation}
f(z'_{t-1}) \approx f(\sqrt{\bar{\alpha}_n} z_{t-1}) + \sqrt{1-\bar{\alpha}_n} \mathbf{J}_f \epsilon + \frac{1-\bar{\alpha}_n}{2} \epsilon^T \mathbf{H}_f \epsilon
\end{equation}
where $\mathbf{H}_f$ represents the Hessian matrix. Substituting this back into the expected loss function, the cross-terms vanish since $\mathbb{E}[\epsilon]=0$, and the loss function transforms into:
\begin{equation}
\mathcal{L}_{noise} \approx \mathcal{L}_{clean} + \underbrace{\frac{1-\bar{\alpha}_n}{2} \text{Tr}(\mathbf{J}_f^T \mathbf{J}_f)}_{\text{Jacobian Regularization}}
\end{equation}
By observing the above equation, we find that adding noise during the training process is equivalent to adding a penalty term for the Frobenius norm of the Jacobian matrix to the objective function. This forces the model to learn a smoother mapping, ensuring that $\rho(\mathbf{J}_f) < 1$. Mathematically, this guarantees the error convergence during recursive generation.

%------------------------------------------------------------------
\section{Detailed theoretical analysis of the scene decoupling estimation module}
\subsection{Conditional Decomposition Theory for Scene Decoupling}
From the perspective of probabilistic modeling, the SDE module decomposes the complex panoramic distribution $P(x_s \mid x_{s-1})$ into two sub-manifolds with distinct motion characteristics. 
The decoupling of the point cloud can be expressed as a decomposition of linear operators:
\begin{equation}
P^{s-1} = \underbrace{M \cdot P^{s-1}}_{P^{s-1}_{obj}} + \underbrace{(1-M) \cdot P^{s-1}}_{P^{s-1}_{bg}}
\end{equation}
Here, $M$ is a mask function defined based on the set of bounding boxes $B^{s-1}$:
\begin{equation}
M(p) = \begin{cases} 1, & \text{if } p \in \bigcup B^{s-1} \\ 0, & \text{otherwise} \end{cases}
\end{equation}
The theoretical basis for this decomposition lies in rigid motion separation: the background follows the global transformation of the static environment, whereas foreground objects adhere to independent local dynamic trajectories.

%------------------------------------------------------------------
\subsection{Foreground Estimation: Residual Mapping Based on Object Center Offsets}
Define $E_{relative}^{s-1} \in SE(3)$ as the ego-motion transformation matrix. First, we perform a viewpoint adjustment:
\begin{equation}
\hat{p}_{ij}^{s-1} = \mathbf{R}_{ego} P_{ij}^{s-1} + \mathbf{t}_{ego}
\end{equation}
This eliminates the visual drift caused by the ego-vehicle's movement.
The deep implication of the estimation equation $\widetilde{p}_{ij}^{s} = \hat{p}_{ij}^{s-1} + (C_{ij}^{s} - C_{ij'}^{s-1})$ is first-order motion compensation. 
Let $\mathbf{v}_{ij}$ be the velocity vector of the $j$-th object of the $i$-th category; then:
\begin{equation}
\Delta C = C_{ij}^{s} - C_{ij'}^{s-1} \approx \int_{s-1}^{s} \mathbf{v}_{ij}(t) dt
\end{equation}
Through this offset, we effectively construct a prediction operator from a Lagrangian perspective. 
It assumes that within an extremely short timestep, the object's geometric deformation is isotropic or invariant:
\begin{equation}
\widetilde{P}^s_{obj} = \mathcal{T}_{\Delta C} (\mathcal{T}_{ego} (P^{s-1}_{obj}))
\end{equation}
where $\mathcal{T}$ represents the translation and rotation operators. 
This approach significantly reduces the generator's burden of predicting the position of dynamic objects, allowing it to focus exclusively on deformation details.

%------------------------------------------------------------------
\subsection{Background Estimation: Simplification of Affine Transformations}
For a background point $p \in P^{s-1}_{bg}$, the complete motion transformation is typically defined as $p' = \mathbf{R}p + \mathbf{t}$. 
However, since the range images are established within the ego-centric LiDAR coordinate system, the aforementioned translational relationship becomes invalid.
Under spherical coordinates or range view mapping, the rotation $\mathbf{R}$ corresponds to a horizontal or vertical shift of the image, whereas the translation $\mathbf{t}$ involves complex pixel-level resampling. 
By simplifying the process to a rotation-only transformation, we preserve the primary consistency features of the background. Meanwhile, the diffusion model is leveraged to learn the variations in sensor observations caused by actual translations.

%------------------------------------------------------------------
\section{Algorithm Procedure and Visualization Results of SDE Module}

%-------------------------------------------------------------------
\begin{algorithm}[htbp]
\caption{Scene Decoupling Estimation}
\label{alg_sde}
\small
\begin{algorithmic}[1]
\STATE \textbf{Input:} Previous LiDAR $ P^{s-1} \in \mathbb{R}^{N^{s-1} \times 4} $ with $ B^{s-1} = \{ b_{ij}^{s-1} \in \mathbb{R}^{8 \times 3}\} $, current $ B^{s} = \{ b_{ij}^{s} \in \mathbb{R}^{8 \times 3}\} $, matrices $ E_{\text{relative}}^{s-1} \in \mathbb{R}^{4 \times 4}$.
\STATE \textbf{Output:} Estimated image of current-frame foreground and background in range-view space $ I^{s}_{\text{obj}}, I^{s}_{\text{bg}} \in \mathbb{R}^{2 \times H \times W} $.

\STATE \textbf{Step 1: Decoupling.} Decouple the scene point cloud into foreground and background based on 3D bounding boxes.
\FOR{$i = 1$ \TO $D$}
    \FOR{$j = 1$ \TO $K_i$}
        \STATE $P_{ij}^{s-1} = \{ p^{s-1} \mid \forall p^{s-1} \text{ in bbox}(b_{ij}^{s-1}) \}$;
    \ENDFOR
\ENDFOR
\STATE $ P_{\text{obj}}^{s-1} = \{ p^{s-1} \in P_{ij}^{s-1} \mid \forall i = 1\TO D, j = 1 \TO K_i \}$;
\STATE $ P_{\text{bg}}^{s-1} = \{ p^{s-1} \mid \forall p^{s-1} \in P^{s-1} \text{ and } p^{s-1} \notin P_{\text{obj}}^{s-1} \}$;

\STATE \textbf{Step 2: Coordinate Transformation.} Perform coordinate transformation on the foreground and background point clouds separately.
\STATE $ \hat{P}_{ij}^{s-1} = \{p_{ij}^{s-1} \cdot E_{\text{rel}}^{s-1}[:3,:3] + E_{\text{rel}}^{s-1}[3,:3] \mid \forall p_{ij}^{s-1} \in P_{ij}^{s-1} \} $;
\STATE $ \widetilde{P}^{s}_{\text{bg}} = \{ p^{s-1}_{\text{bg}} \cdot E_{\text{rel}}^{s-1}[:3,:3] \mid \forall p^{s-1} \in P_{\text{bg}}^{s-1} \} $;

\STATE \textbf{Step 3: Category Search.} Search for nearest bounding box in center by analogy and match.
\FOR{$i = 1$ \TO $D$}
    \FOR{$j = 1$ \TO $K_i$}
        \STATE $ c_{ij}^{s-1} = \text{Center}(b_{ij}^{s-1})$;
        \STATE $ c_{ij}^{s} = \text{Center}(b_{ij}^{s})$;
        \STATE $ c_{ij'}^{s-1} = \text{Search\_near}(c_{ij}^{s-1}, \{c_{ij}^{s} \mid \forall i,j \}) $;
    \ENDFOR
\ENDFOR

\STATE \textbf{Step 4: Match.} Match the object from the previous frame to the current scene.
\STATE $ \widetilde{P}^{s}_{ij} = \hat{P}_{ij}^{s-1} + (c_{ij}^{s} - c_{ij'}^{s-1}) , \forall i = 1$ \TO $D, j = 1$ \TO $K_i $;

\STATE \textbf{Step 5: Merge.} Merge foreground object point clouds of all analogies.
\STATE $ \widetilde{P}^{s}_{\text{obj}} = \{ p^{s} \in \widetilde{P}^{s}_{ij} \mid \forall i = 1$ \TO $D, j = 1$ \TO $K_i \}$;

\STATE \textbf{Spherical Projection.} Project both $\widetilde{P}^{s}_{\text{obj}}$ and $\widetilde{P}^{s}_{\text{bg}}$ into range-view space.
\STATE $ \widetilde{I}^{s}_{\text{obj}} = \text{Sph\_proj} (\widetilde{P}^{s}_{\text{obj}}) $;
\STATE $ \widetilde{I}^{s}_{\text{bg}} = \text{Sph\_proj} (\widetilde{P}^{s}_{\text{bg}}) $;

\STATE \textbf{Return:} Estimated current-frame foreground and background point clouds $\widetilde{P}^{s}_{\text{obj}}$ and $\widetilde{P}^{s}_{\text{bg}}$ with their range image $ I^{s}_{\text{obj}},  I^{s}_{\text{bg}} \in \mathbb{R}^{2 \times H \times W} $.
\end{algorithmic}
\end{algorithm}
%------------------------------------------------------------------

\begin{figure}[t]
  \centering
  \includegraphics[width=0.7\linewidth]{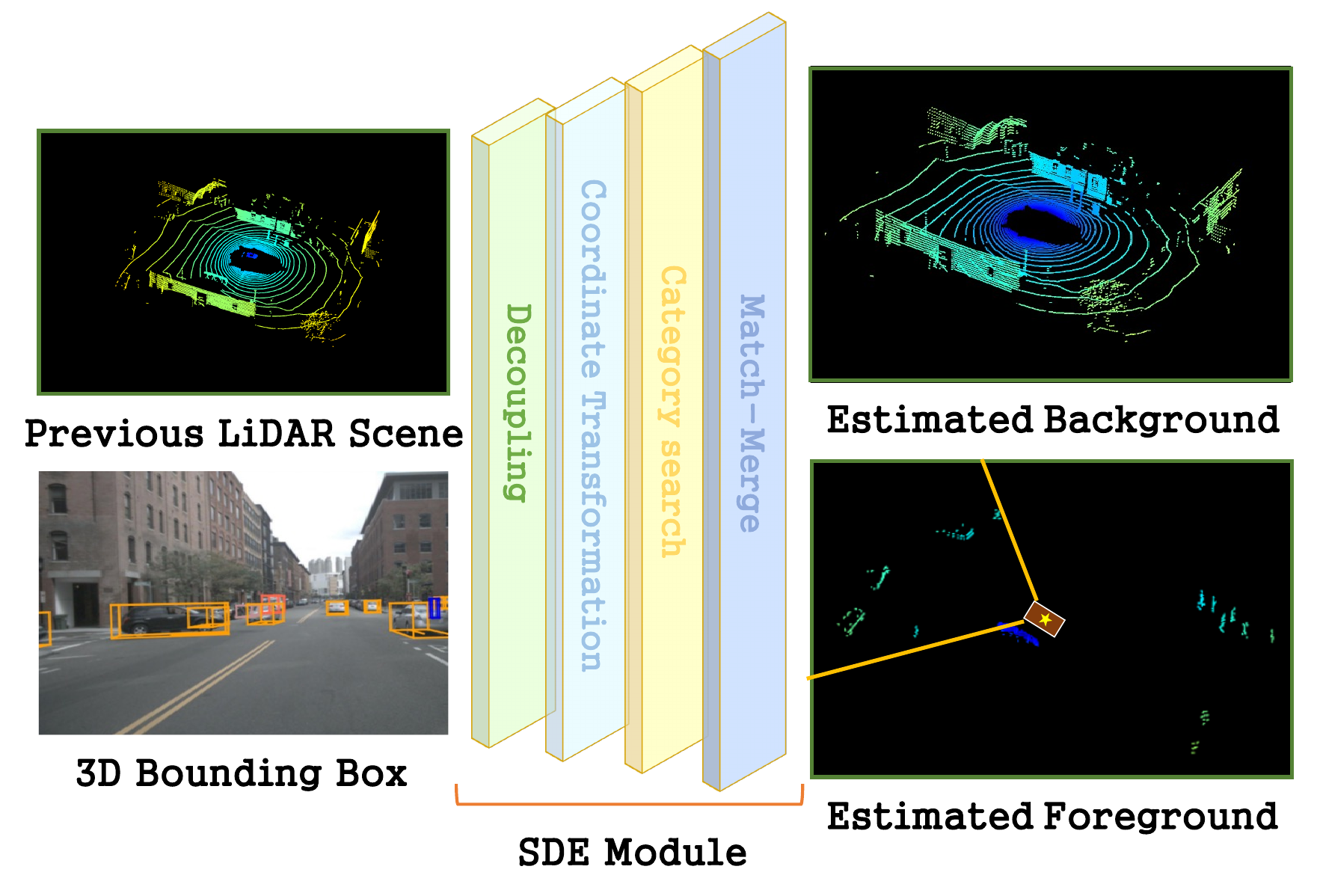}\\
  \caption{\label{sde_result}
    Visualization of real input and output of SDE module.}
\end{figure}

In Algorithm Flowchart ~\ref{alg_sde}, we demonstrate the detailed execution process of the SDE module.
In Figure~\ref{sde_result}, we present the input and corresponding output of the SDE module for a specific driving scene. 
It should be noted that, to facilitate a better understanding of the shape and position of bounding boxes in the real world, we mark them in the camera-captured images (see the lower left corner of Figure~\ref{sde_result}).
Due to the limited field of view of the camera, we select a frontal perspective and illustrate the viewpoint within the foreground estimated by the SDE module (see the lower right corner of Figure~\ref{sde_result}).
These visualizations demonstrate that LaGen consistently achieves superior performance across three metrics for each frame.

%------------------------------------------------------------------
%------------------------------------------------------------------

%--------------------------------------------------------------
\section{Robustness Experiments on 3D Bounding Box Conditioning}
To further evaluate the robustness of the LaGen framework, we conducted additional experiments by perturbing the 3D bounding box conditioning, including random shifting and random dropping. 
Specifically, each bounding box was randomly shifted with a probability of 0.5, where the shift magnitude was set to 1 m, 2 m, 4 m, and 8 m, respectively. For random dropping, the dropping probability was set to 0.1, 0.2, 0.3, and 0.5. 
The detailed experimental results are reported in Table~\ref{table_ablation_bbox}.

%--------------------------------------------------------------
\begin{table}[ht]
\centering
\caption{The results of robustness experiments on 3D bounding box conditioning. MMD values are reported in $10^{-4}$ and JSD in $10^{-2}$.}
\label{table_ablation_bbox}
\renewcommand{\arraystretch}{1.1}
\resizebox{0.5\columnwidth}{!}{
\small
\begin{tabular}{c|l|cc} 
\toprule
\textbf{Mode} & \textbf{Configuration}  & \textbf{MMD} & \textbf{JSD} \\
\midrule 
\midrule 
\rowcolor{myyellow}
\multirow{1}{*}{Standard} 
& \qquad \quad \#  & \textbf{0.109} & \textbf{2.77} \\
\midrule 
\multirow{4}{*}{\makecell{Random \\ Perturbation}} 
& [-1.0m, +1.0m]     & 0.190 & 3.63 \\
& [-2.0m, +2.0m]     & 0.221 & 3.70 \\
& [-4.0m, +4.0m]     & 0.310 & 3.90  \\
& [-8.0m, +8.0m]     & 0.335 & 3.97  \\
\midrule 
\multirow{4}{*}{\makecell{Random \\ Drop}} 
& \qquad 0.10  & 0.207 & 3.54  \\
& \qquad 0.20  & 0.256 & 3.68  \\
& \qquad 0.30  & 0.340 & 3.87  \\
& \qquad 0.50  & 0.463 & 4.23  \\
\bottomrule
\end{tabular}
}
\end{table}
%--------------------------------------------------------------

It can be observed that our proposed framework maintains high-fidelity generation even under unstable bounding box information (retaining a certain advantage over other state-of-the-art generation methods).

%--------------------------------------------------------------
\section{More Quantitative Results}

\begin{figure}[t]
  \centering
  \includegraphics[width=\linewidth]{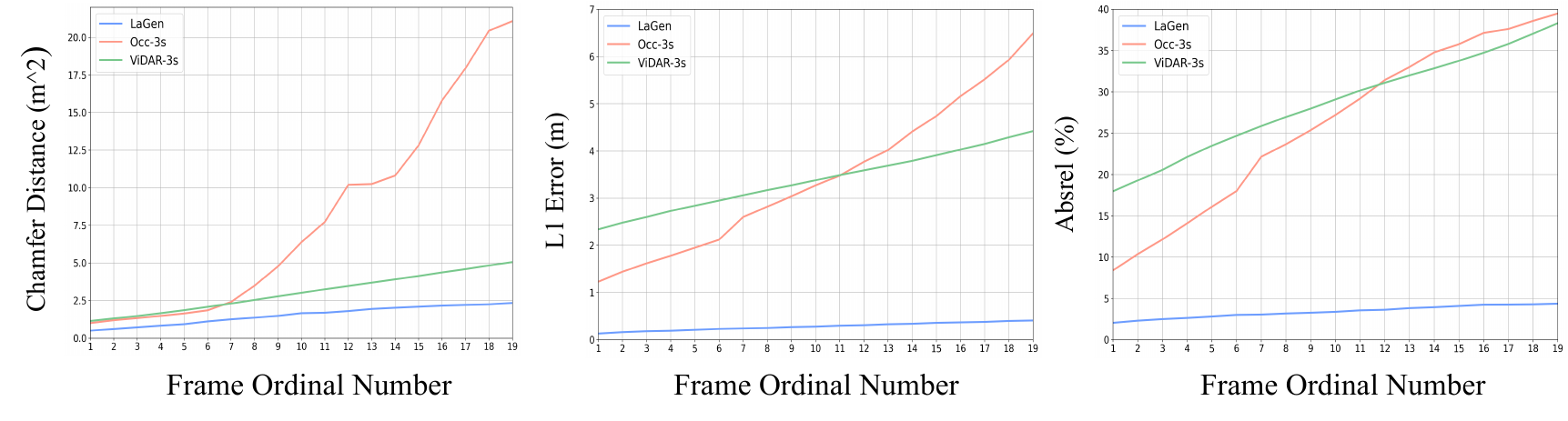}\\
  \caption{\label{l1}
    The cd, L1 error metirc and absrel metirc curve of LaGen and baseline methods.}
\end{figure}

In this section, we present detailed frame-by-frame comparison results between LaGen and two other state-of-the-art prediction-based approaches, as shown in Tables \ref{table_extra_metric1}, \ref{table_extra_metric2}, \ref{table_extra_metric3}, and \ref{table_extra_metric4}.
We can clearly see that LaGen achieves optimal performance in every frame of the three metrics (see Figure~\ref{l1}).

Why is LaGen able to outperform these prediction frameworks that require multiple frames of historical input, even though it only takes a single-frame LiDAR scene as input? 
We conducted an in-depth analysis of the essence of these prediction-based frameworks. 
In practice, they mainly learn the ego-vehicle motion and scene dynamics within a short time span and then apply this learned pattern to subsequent frames.
This approach means that they cannot accurately predict scene information over long time spans. 
In real driving scenarios, the behavior of the ego-vehicle and surrounding objects can exhibit significant variability, making it unrealistic for them to always follow consistent dynamics.

In contrast, our proposed framework generates scenes frame by frame and is able to fully leverage additional scene information to facilitate the generation of future LiDAR scenes. 
For each frame, it can analyze object-state changes using 3D bounding box information, resulting in more realistic and reasonable scene layouts.
Moreover, an interactive generation framework can not only faithfully depict high-fidelity LiDAR scenes, but also exhibit diversity by generating many possible futures.

%------------------------------------------------------------------
%--------------------------------------------------------------

%--------------------------------------------------------------
\section{More Qualitative Results}

\begin{figure*}[ht]
  \centering
  \includegraphics[width=1\linewidth]{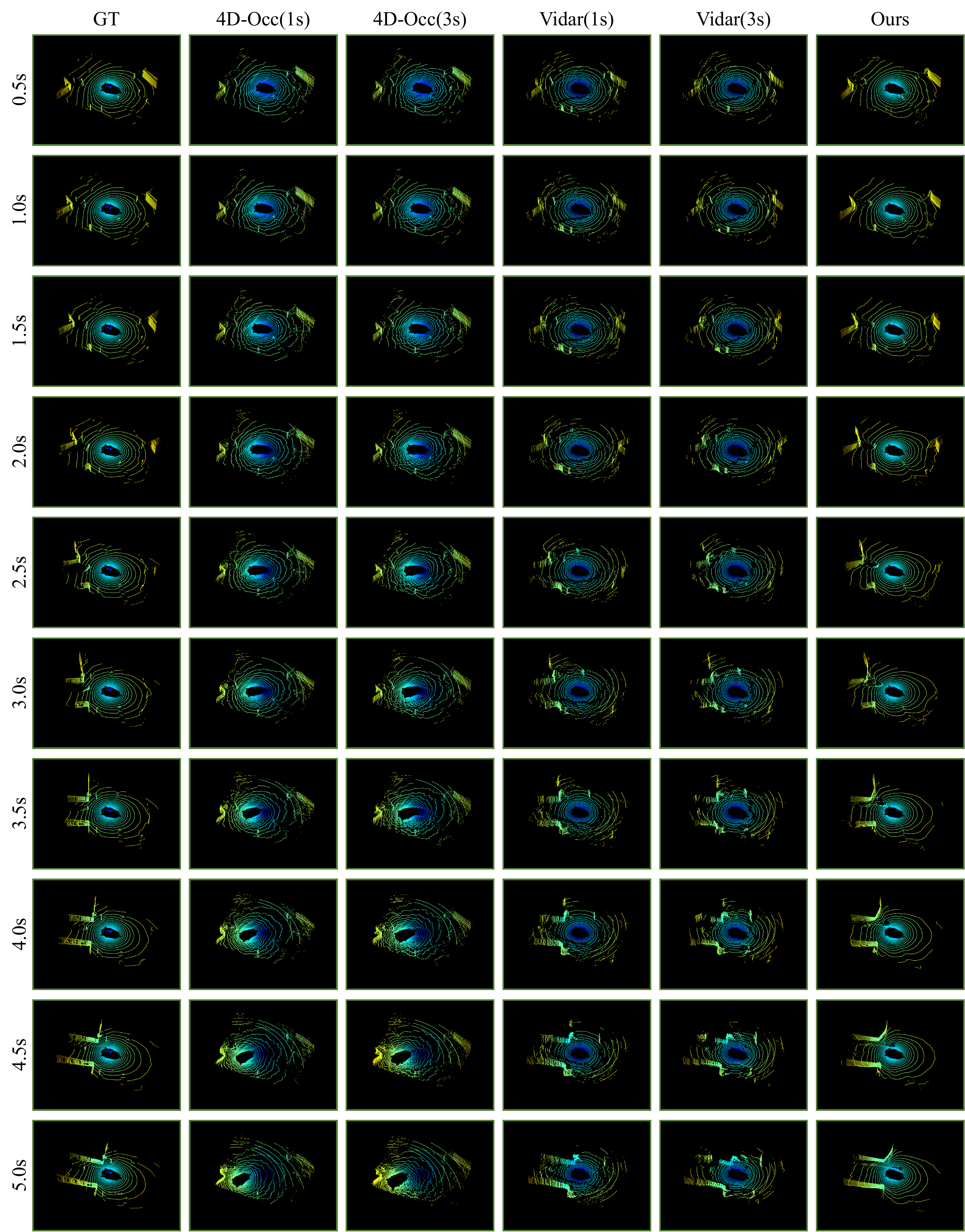}\\
  \caption{\label{extra_scenes11}
    Additional qualitative comparison between LaGen and two other advanced prediction methods in a specific scenario. We present visualizations of LiDAR scenes for all frames within the 0.5s–9.5s interval, allowing detailed observation of ego-vehicle motion and scene evolution trends (this part displays the time period from 0.5s to 5.0s). In this scenario, the ego-vehicle first turns slightly to the left and then proceeds straight along the road.}
  \end{figure*}

\begin{figure*}[ht]
  \centering
  \includegraphics[width=1\linewidth]{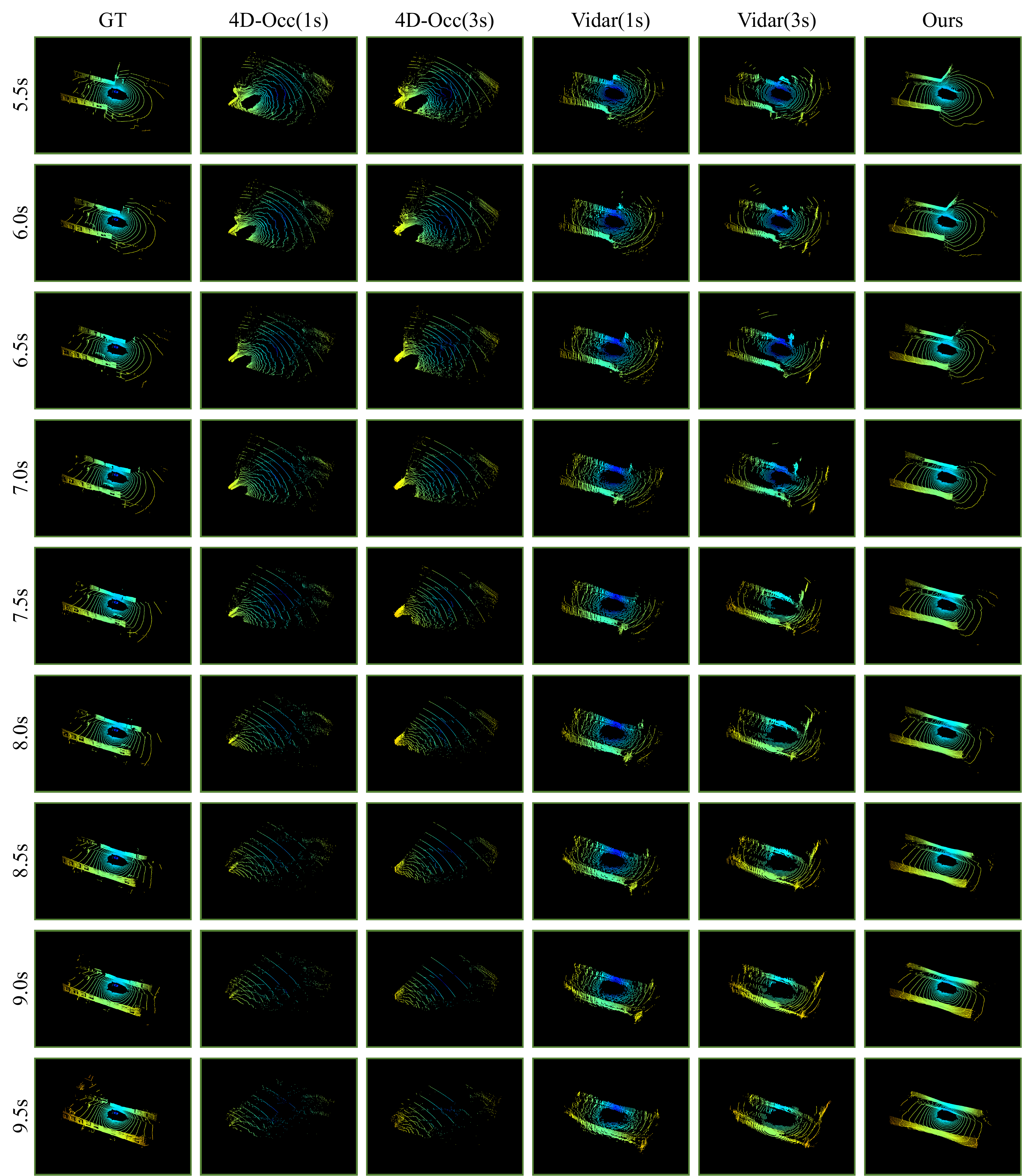}\\
  \caption{\label{extra_scenes12}
    Additional qualitative comparison between LaGen and two other advanced prediction methods in a specific scenario. We present visualizations of LiDAR scenes for all frames within the 0.5s–9.5s interval, allowing detailed observation of ego-vehicle motion and scene evolution trends (this part displays the time period from 5.5s to 9.5s). In this scenario, the ego-vehicle first turns slightly to the left and then proceeds straight along the road.}
  \end{figure*}

\begin{figure*}[ht]
  \centering
  \includegraphics[width=1\linewidth]{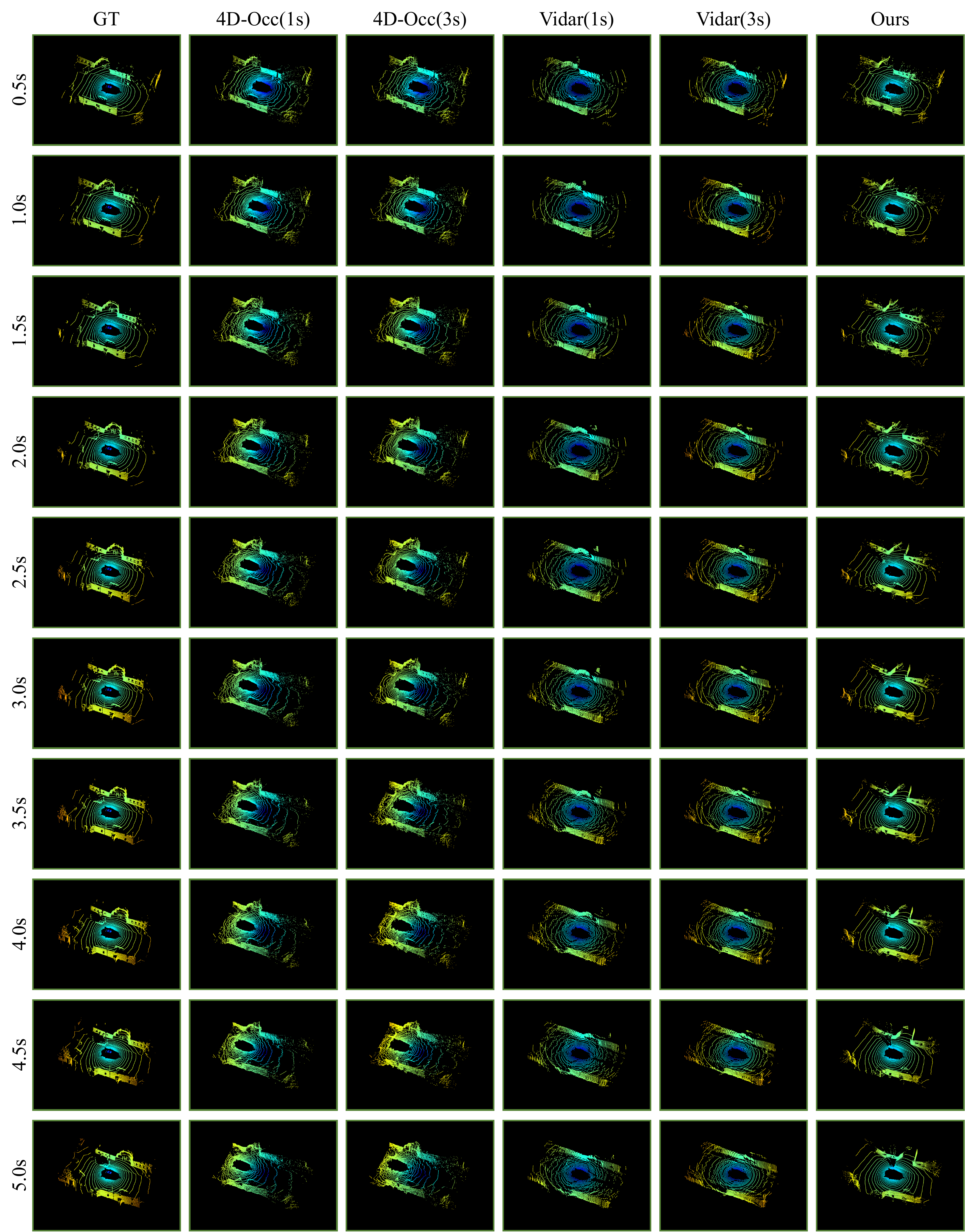}\\
  \caption{\label{extra_scenes21}
    Additional qualitative comparison between LaGen and two other advanced prediction methods on another scenario. We present visualizations of LiDAR scenes for all frames within the 0.5s–9.5s interval, allowing detailed observation of ego-vehicle motion and scene evolution trends (this part displays the time period from 0.5s to 5.0s). In this scenario, the ego-vehicle is slowly moving straight ahead.}
  \end{figure*}

\begin{figure*}[ht]
  \centering
  \includegraphics[width=1\linewidth]{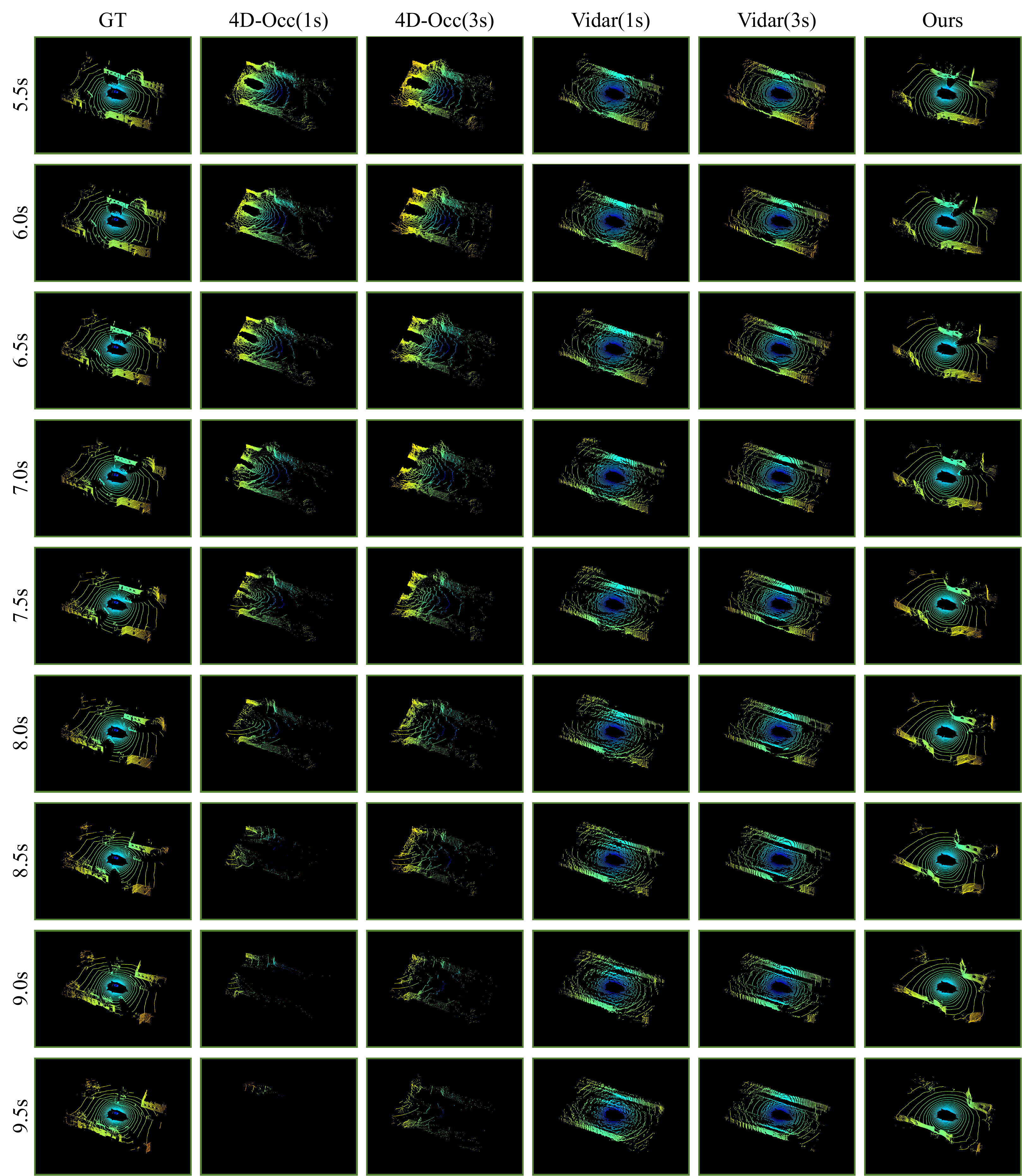}\\
  \caption{\label{extra_scenes22}
    Additional qualitative comparison between LaGen and two other advanced prediction methods on another scenario. We present visualizations of LiDAR scenes for all frames within the 0.5s–9.5s interval, allowing detailed observation of ego-vehicle motion and scene evolution trends (this part displays the time period from 5.5s to 9.5s). In this scenario, the ego-vehicle is slowly moving straight ahead.}
  \end{figure*}

In this section, we provide qualitative frame-by-frame comparison results between LaGen and two prediction-based baseline methods, as illustrated in Figure \ref{extra_scenes11}, \ref{extra_scenes12} and Figure \ref{extra_scenes21}, \ref{extra_scenes22}.

We highlight two representative driving scenarios for comparison.
In the first scenario, the self driving vehicle first makes a left turn and then continues straight after entering the new road.
In the second scenario, the ego-vehicle maintains a slow, straight trajectory throughout.
By presenting frame-by-frame visualizations for these driving scenarios, we can more clearly observe the behavior of the ego-vehicle and the evolution of the surrounding scene.

%------------------------------------------------------------------

\newpage

%--------------------------------------------------------------
\setlength{\tabcolsep}{2.8pt} 
\begin{table*}[p]
\centering
\caption{Quantitative comparison results of LaGen and two other advanced prediction methods over the time interval of 0.5s to 2.5s.}
\renewcommand{\arraystretch}{1.15}
\resizebox{\textwidth}{!}{ 
\begin{tabular}{ccccccc|ccccc|ccccc}
\toprule
 \textbf{Method} & \textbf{Input} &
\multicolumn{15}{c}{\textbf{Chamfer Distance (m$^2$)$\downarrow$} \qquad \qquad \enspace \textbf{L1 Error (m)$\downarrow$} \qquad \qquad \qquad \qquad \textbf{Absrel (\%)$\downarrow$}} \qquad \quad\\
\cmidrule(l{3pt}r{3pt}){3-17} 
&\textbf{frames}& 0.5s & 1.0s & 1.5s & 2.0s & 2.5s &0.5s & 1.0s & 1.5s & 2.0s & 2.5s  &0.5s & 1.0s & 1.5s & 2.0s & 2.5s \\
\midrule
4D-Occ
& 2 
& 1.15 & 1.61 & \underline{1.84} & \underline{2.15} & \underline{2.56} & 1.24 & 1.55 & \underline{2.04} & \underline{2.28} & \underline{2.72} & 8.88 & 11.79 & \underline{14.82} & \underline{16.85} & \underline{19.57} \\
4D-Occ  
& 6 
& \cellcolor{Lavender} 1.00 & \cellcolor{Lavender} 1.19 & \cellcolor{Lavender} 1.33 & \cellcolor{Lavender} 1.48 & \cellcolor{Lavender} 1.64 & \cellcolor{Lavender} 1.23 & \cellcolor{Lavender} 1.44 & \cellcolor{Lavender} 1.62 & \cellcolor{Lavender} 1.78 & \cellcolor{Lavender} 1.95 & \cellcolor{Lavender} 8.41 & \cellcolor{Lavender} 10.37 & \cellcolor{Lavender} 12.15 & \cellcolor{Lavender} 14.10 & \cellcolor{Lavender} 16.11 \\
\midrule
ViDAR 
& 2 
& 1.15 & 1.36 & 1.56 & 1.75 & 1.98 &  2.38 & 2.55 & 2.70 & 2.85 & 3.00 & 16.29 & 18.09 & 19.96 & 21.82 & 23.73 \\
ViDAR 
& 6
& 1.14 & 1.31 & 1.47 & 1.66 & 1.86 & 2.34 & 2.48 & 2.60 & 2.73 & 2.84 & 17.98 & 19.29 & 20.57 & 22.13 & 23.49 \\
\midrule
LaGen & 
\cellcolor{rowgray}1
& \cellcolor{myred} \textbf{0.50} & \cellcolor{myred} \textbf{0.61} & \cellcolor{myred} \textbf{0.72} & \cellcolor{myred} \textbf{0.83} & \cellcolor{myred} \textbf{0.926} & \cellcolor{myred} \textbf{0.13}  & \cellcolor{myred} \textbf{0.16} & \cellcolor{myred} \textbf{0.18} & \cellcolor{myred} \textbf{0.20} & \cellcolor{myred} \textbf{0.21} & \cellcolor{myred} \textbf{2.04} & \cellcolor{myred} \textbf{2.31} & \cellcolor{myred} \textbf{2.47} & \cellcolor{myred} \textbf{2.72} & \cellcolor{myred} \textbf{2.89}  \\
\bottomrule
\end{tabular}
\label{table_extra_metric1}
}
\end{table*}
%--------------------------------------------------------------

%--------------------------------------------------------------
\setlength{\tabcolsep}{2.8pt} 
\begin{table*}[p]
\centering
\caption{Quantitative comparison results of LaGen and two other advanced prediction methods over the time interval of 3.0s to 5.0s.}
\renewcommand{\arraystretch}{1.15}
\resizebox{\textwidth}{!}{ 
\begin{tabular}{ccccccc|ccccc|ccccc}
\toprule
 \textbf{Method} & \textbf{Input} &
\multicolumn{15}{c}{\textbf{Chamfer Distance (m$^2$)$\downarrow$} \qquad \qquad \enspace \textbf{L1 Error (m)$\downarrow$} \qquad \qquad \qquad \qquad \textbf{Absrel (\%)$\downarrow$}} \qquad \quad\\
\cmidrule(l{3pt}r{3pt}){3-17} 
&\textbf{frames}& 3.0s & 3.5s & 4.0s & 4.5s & 5.0s & 3.0s & 3.5s & 4.0s & 4.5s & 5.0s & 3.0s  & 3.5s & 4.0s & 4.5s & 5.0s \\
\midrule
4D-Occ 
& 2 
& \underline{3.10} & \underline{4.30} & \underline{6.24} & \underline{7.51} & \underline{8.72} & \underline{2.97} & \underline{3.41} & \underline{3.72} & \underline{4.16} & \underline{4.37} & \underline{21.84} & \underline{25.36} & \underline{28.26} & \underline{32.27} & \underline{34.77} \\
4D-Occ  
& 6 
& \cellcolor{Lavender} 1.85 & \underline{2.41} & \underline{3.49} & \underline{4.77} & \underline{6.38} & \cellcolor{Lavender} 2.12 & \cellcolor{Lavender} \underline{2.60} & \cellcolor{Lavender} \underline{2.82} & \cellcolor{Lavender} \underline{3.04} & \cellcolor{Lavender} \underline{3.27} & \cellcolor{Lavender} 18.01 & \cellcolor{Lavender} \underline{22.16} & \cellcolor{Lavender} \underline{23.68} & \cellcolor{Lavender} \underline{25.37} & \cellcolor{Lavender} \underline{27.22} \\
\midrule
ViDAR 
& 2 
& 2.22 & 2.50 & 2.81 & 3.11 & 3.45 & 3.16 & 3.34 & 3.54 & 3.74 & 3.97 & 25.48 & 27.34 & 29.23 & 31.15 & 33.41 \\
ViDAR 
& 6
& 2.08 & \cellcolor{Lavender} 2.30 & \cellcolor{Lavender} 2.54 & \cellcolor{Lavender} 2.78 & \cellcolor{Lavender} 3.01 & 2.95 & 3.06 & 3.17 & 3.27 & 3.38 & 24.70 & 25.87 & 26.97 & 27.99 & 29.09 \\
\midrule
LaGen & 
\cellcolor{rowgray}1
& \cellcolor{myred} \textbf{1.11} & \cellcolor{myred} \textbf{1.25} & \cellcolor{myred} \textbf{1.36} & \cellcolor{myred} \textbf{1.48} & \cellcolor{myred} \textbf{1.66} & \cellcolor{myred} \textbf{0.23}  & \cellcolor{myred} \textbf{0.24} & \cellcolor{myred} \textbf{0.26} & \cellcolor{myred} \textbf{0.26} & \cellcolor{myred} \textbf{0.28} & \cellcolor{myred} \textbf{2.98} & \cellcolor{myred} \textbf{3.03} & \cellcolor{myred} \textbf{3.21} & \cellcolor{myred} \textbf{3.30} & \cellcolor{myred} \textbf{3.35}  \\
\bottomrule
\end{tabular}
\label{table_extra_metric2}
}
\end{table*}
%--------------------------------------------------------------

%--------------------------------------------------------------
\setlength{\tabcolsep}{2.5pt} 
\begin{table*}[p]
\centering
\caption{Quantitative comparison results of LaGen and two other advanced prediction methods over the time interval of 5.5s to 7.5s.}
\renewcommand{\arraystretch}{1.15}
\resizebox{\textwidth}{!}{ 
\begin{tabular}{ccccccc|ccccc|ccccc}
\toprule
 \textbf{Method} & \textbf{Input} &
\multicolumn{15}{c}{\textbf{Chamfer Distance (m$^2$)$\downarrow$} \qquad \qquad \quad \textbf{L1 Error (m)$\downarrow$} \qquad \qquad \qquad \qquad \textbf{Absrel (\%)$\downarrow$}} \quad \\
\cmidrule(l{3pt}r{3pt}){3-17} 
&\textbf{frames}& 5.5s & 6.0s & 6.5s & 7.0s & 7.5s & 5.5s & 6.0s & 6.5s & 7.0s & 7.5s & 5.5s & 6.0s & 6.5s & 7.0s & 7.5s  \\
\midrule
4D-Occ 
& 2 
& \underline{10.20} & \underline{12.31} & \underline{16.74} & \underline{20.35} & \underline{25.11} & \underline{4.87} & \underline{5.11} & \underline{5.89} & \underline{6.33} & \underline{7.07} & \underline{40.14} & \underline{42.60} & \underline{48.79} & \underline{50.68} & \underline{56.24} \\
4D-Occ 
& 6 
& \underline{7.73} & \underline{10.20} & \underline{10.25} & \underline{10.81} & \underline{12.82} & \cellcolor{Lavender} \underline{3.48} & \underline{3.77} & \underline{4.02} & \underline{4.41} & \underline{4.74} & \cellcolor{Lavender} \underline{29.20} & \underline{31.45} & \underline{33.02} & \underline{34.80} & \underline{35.78} \\
\midrule
ViDAR 
& 2 
& 3.79 & 4.15 & 4.51 & 4.85 & 5.18 & 4.25 & 4.56 & 4.87 & 5.15 & 5.40 & 36.25 & 39.59 & 42.99 & 46.03 & 48.61 \\
ViDAR 
& 6
& \cellcolor{Lavender} 3.25 & \cellcolor{Lavender} 3.47 & \cellcolor{Lavender} 3.69 & \cellcolor{Lavender} 3.91 & \cellcolor{Lavender} 4.13 & 3.49 & \cellcolor{Lavender} 3.59 & \cellcolor{Lavender} 3.69 & \cellcolor{Lavender} 3.79 & \cellcolor{Lavender} 3.91 & 30.18 & \cellcolor{Lavender} 31.12 & \cellcolor{Lavender} 32.01 & \cellcolor{Lavender} 32.86 & \cellcolor{Lavender} 33.77 \\
\midrule
LaGen & 
\cellcolor{rowgray}1
& \cellcolor{myred} \textbf{1.69} & \cellcolor{myred} \textbf{1.80} & \cellcolor{myred} \textbf{1.94} & \cellcolor{myred} \textbf{2.02} & \cellcolor{myred} \textbf{2.09} & \cellcolor{myred} \textbf{0.28}  & \cellcolor{myred} \textbf{0.28} & \cellcolor{myred} \textbf{0.29} & \cellcolor{myred} \textbf{0.29} & \cellcolor{myred} \textbf{0.30} & \cellcolor{myred} \textbf{3.40} & \cellcolor{myred} \textbf{3.38} & \cellcolor{myred} \textbf{3.34} & \cellcolor{myred} \textbf{3.48} & \cellcolor{myred} \textbf{3.63}\\
\bottomrule
\end{tabular}
\label{table_extra_metric3}
}
\end{table*}
%--------------------------------------------------------------

%--------------------------------------------------------------
\setlength{\tabcolsep}{5.3pt} 
\begin{table*}[p]
\centering
\caption{Quantitative comparison results of LaGen and two other advanced prediction methods over the time interval of 8.0s to 9.5s.}
\renewcommand{\arraystretch}{1.15}
\resizebox{\textwidth}{!}{ 
\begin{tabular}{cccccc|cccc|cccc}
\toprule
 \textbf{Method} & \textbf{Input} &
\multicolumn{12}{c}{\textbf{Chamfer Distance (m$^2$)$\downarrow$} \qquad \qquad \enspace \textbf{L1 Error (m)$\downarrow$} \qquad \qquad \qquad \quad \textbf{Absrel (\%)$\downarrow$}} \qquad  \\
\cmidrule(l{3pt}r{3pt}){3-14} 
&\textbf{frames}& 8.0s & 8.5s & 9.0s & 9.5s & 8.0s & 8.5s & 9.0s & 9.5s & 8.0s & 8.5s & 9.0s & 9.5s  \\
\midrule
4D-Occ 
& 2 
& \underline{35.67} & \underline{42.60} & \underline{55.47} & \underline{65.92} & \underline{8.39} & \underline{9.00} & \underline{9.95} & \underline{10.62} & \underline{62.50} & \underline{63.76} & \underline{67.37} & \underline{67.82} \\
4D-Occ  
& 6 
& \underline{15.81} & \underline{17.95} & \underline{20.44} & \underline{21.08} & \underline{5.16} & \underline{5.52} & \underline{5.93} & \underline{6.49} & \underline{37.14} & \underline{37.63} & \underline{38.61} & \underline{39.50} \\
\midrule
ViDAR 
& 2 
& 5.49 & 5.79 & 6.09 & 6.37 & 5.62 & 5.81 & 5.95 & 6.06 & 50.68 & 52.31 & 53.44 & 54.13 \\
ViDAR 
& 6
& \cellcolor{Lavender} 4.37 & \cellcolor{Lavender} 4.59 & \cellcolor{Lavender} 4.83 & \cellcolor{Lavender} 5.06 & \cellcolor{Lavender} 4.03 & \cellcolor{Lavender} 4.15 & \cellcolor{Lavender} 4.29 & \cellcolor{Lavender} 4.42 & \cellcolor{Lavender} 34.75 & \cellcolor{Lavender} 35.81 & \cellcolor{Lavender} 37.05 & \cellcolor{Lavender} 38.34 \\
\midrule
LaGen & 
\cellcolor{rowgray}1
& \cellcolor{myred} \textbf{2.16} & \cellcolor{myred} \textbf{2.21} & \cellcolor{myred} \textbf{2.25} & \cellcolor{myred} \textbf{2.34} & \cellcolor{myred} \textbf{0.30} & \cellcolor{myred} \textbf{0.31} & \cellcolor{myred} \textbf{0.31} & \cellcolor{myred} \textbf{0.32} &  \cellcolor{myred} \textbf{3.56} & \cellcolor{myred} \textbf{3.54} & \cellcolor{myred} \textbf{3.60} & \cellcolor{myred} \textbf{3.61} \\
\bottomrule
\end{tabular}
\label{table_extra_metric4}
}
\end{table*}
%--------------------------------------------------------------

%--------------------------------------------------------------------
%-------------------------------------------------------------------

%--------------------------------------------------------------------
%-------------------------------------------------------------------

%-------------------------------------------------------------------
%-------------------------------------------------------------------

%-------------------------------------------------------------------
%-------------------------------------------------------------------
%%%%%%%%%%%%%%%%%%%%%%%%%%%%%%%%%%%%%%%%%%%%%%%%%%%%%%%%%%%%%%%%%%%%%%%%%%%%%%%
%%%%%%%%%%%%%%%%%%%%%%%%%%%%%%%%%%%%%%%%%%%%%%%%%%%%%%%%%%%%%%%%%%%%%%%%%%%%%%%
%\fi

\end{document}